\documentclass{article}

\usepackage[utf8]{inputenc} 
\usepackage[T1]{fontenc}    
\usepackage[hidelinks]{hyperref}       
\usepackage{url}            
\usepackage{booktabs}       
\usepackage{amsfonts}       
\usepackage{nicefrac}       
\usepackage{microtype}      

\usepackage{amsmath}
\usepackage{threeparttable}
\usepackage{algorithm}
\usepackage{algpseudocode}
\usepackage{mathtools}
\usepackage{graphicx}
\usepackage{enumitem}
\usepackage{natbib}

\usepackage[accepted]{icml2019}

\newcommand{\argmax}{\mathop{\mathrm{argmax}}}
\newcommand{\expect}{\mathop{\mathbb{E}}}

\newcommand{\fig}[1]{Figure\;\ref{fig:#1}}

\graphicspath{ {img/} }

\begin{document}


\twocolumn[
\icmltitle{Classification with Costly Features as a Sequential Decision-Making Problem}

\icmlsetsymbol{equal}{*}

\begin{icmlauthorlist}
\icmlauthor{Jaromír Janisch}{cvut}
\icmlauthor{Tomáš Pevný}{cvut}
\icmlauthor{Viliam Lisý}{cvut}
\end{icmlauthorlist}

\icmlaffiliation{cvut}{Artificial Intelligence Center, Department of Computer Science, Faculty of Electrical Engineering, Czech Technical University in Prague}

\icmlcorrespondingauthor{Jaromír Janisch}{jaromir.janisch@fel.cvut.cz}


\vskip 0.3in
]

\printAffiliationsAndNotice{}

\begin{abstract}
This work focuses on a specific classification problem, where the information about a sample is not readily available, but has to be acquired for a cost, and there is a per-sample budget. Inspired by real-world use-cases, we analyze \emph{average} and \emph{hard} variations of a \emph{directly specified} budget. We postulate the problem in its explicit formulation and then convert it into an equivalent MDP, that can be solved with deep reinforcement learning. Also, we evaluate a real-world inspired setting with sparse training dataset with missing features. The presented method performs robustly well in all settings across several distinct datasets, outperforming other prior-art algorithms. The method is flexible, as showcased with all mentioned modifications and can be improved with any domain independent advancement in RL.

\end{abstract}

\section{Introduction} \label{sec:introduction}
Classification with Costly Features (CwCF) is a family of classification problems with a cost of acquiring information. This cost can appear in many forms. Usually, it is about money or time, but it is present in any domain with limited resources. We view the problem as a sequential decision-making problem. At each step, based on the information acquired so far, the algorithm has to decide whether to acquire another piece of information (a \emph{feature}) or to classify.

Think about a doctor who is about to make a diagnosis for their patient. There is a number of examinations, tests or analysis which can be made, but each of them has a cost. As much as the doctor wants to make a reliable prediction, he is bound by his \emph{average budget} he should not exceed. Naturally, patients with complicated diseases require more complicated and expensive tests, while trivial problems can be diagnosed with much fewer resources. 

As another motivating example, imagine an online service that analyzes computer files potentially infected with malware. The service is bound by a service-level agreement and has to provide a decision in a specified time, and this time cannot be exceeded. This is an example of a \emph{hard budget}. The process can analyze the files in multiple ways and compute their features, and each computation takes a different, but known amount of time. The goal is to provide accurate predictions, while not violating the time constraint.

In other domains, different requirements arise. One domain contains a lot of missing data, other has imbalanced datasets. There can be a requirement of incorporating an existing classifier into the process. Misclassification errors can have outcomes with different impact, measured in the amount of lost resources.
As we see, there are many variants of the CwCF problem. Techniques adapted for specific problems exist and it is difficult to modify these methods. In this article, we present a flexible reinforcement-learning based framework, that can work with all the mentioned instances. Should other needs arise, the method is easily modified to suit the problem. We mainly demonstrate our method in cases of average and hard budgets and also with missing features. 

The power and generality of our method arise from the decoupling of the problem and the method itself. By using a general reinforcement learning algorithm, we are able to modify the problem specification, and the method will still provide a good result. The core of our algorithm is built on optimal methods, but we lose the guarantees by using function approximation (specifically, neural networks). Our method is also robust to hyperparameter selection, where the same set of hyperparameters usually works well across different domains and settings.

To the best of our knowledge, the presented method is first that can work with both average and hard budgets, is flexible and robust. Formerly, CwCF with the average budget was approached with linear programming \citep{wang2014model}, tree-based algorithms \citep{nan2016pruning}, gradient-based methods \citep{contardo2016recurrent} and recently reinforcement learning \citep{shim2018joint,janisch2019classification}. There are also several publications focusing on the hard budget problem -- guided selection using a heuristic \citep{kapoor2005learning}, and theoretical analyses \citep{cesa2011efficient,zolghadr2013online}. We present an overview of the related work in Section\;\ref{sec:related_work}.

We directly build on recent work of \citet{janisch2019classification}, where the authors established a new state-of-the-art method in the average budget case. The authors proposed to solve the problem of minimizing expected classification loss, along with $\lambda$-scaled total cost:
\begin{equation}\label{eq:problem_original}  
  \min_\theta \expect_{(x,y) \in \mathcal D} \big[ \ell(y_\theta(x), y) + \lambda z_\theta(x) \big]
\end{equation}
where $(x,y)$ are samples taken from the dataset $\mathcal D$, $\ell$ is a classification loss, $\lambda$ is a trade-off parameter, $y_\theta$ is the classifier and $z_\theta$ returns the total cost of used features in the classification.

The definition only focuses on the average budget problem and it also introduces an unintuitive parameter $\lambda$. The user has no option but to \emph{try} different values and see whether the learned model corresponds to a targeted budget or not. In this work we take a step back and propose the definition of the problem in its natural form. Given an explicit budget $b$, the problem for the average case is:
\begin{equation}\label{eq:problem_average}  
  \min_\theta \expect \big[\ell(y_\theta, y) \big], \quad
  \text{s.t. } \expect \big[ z_\theta(x) \big] \leq b
\end{equation}
where the expectations are w.r.t. the distribution of the samples in the dataset. For the hard budget it is:
\begin{equation}\label{eq:problem_hard}  
  \min_\theta \expect \big[\ell(y_\theta, y) \big], \quad
  \text{s.t. } z_\theta(x) \leq b, \forall x 
\end{equation}

\citet{dulac2011datum} showed that the minimization problem \eqref{eq:problem_original} could be transformed into an MDP formulation and solved through standard reinforcement learning techniques. The approach was later improved by \citet{janisch2019classification} with deep learning. We follow this approach and modify the algorithm for both \eqref{eq:problem_average} and \eqref{eq:problem_hard}. In the case of the average budget, we convert the definition \eqref{eq:problem_average} into a Lagrangian framework and search for the optimum with a gradient-based method. In the case of a hard budget, we show that simple modification to the MDP definition enforces the hard constraints. In exact settings, the method would yield an optimal solution and it only lacks guarantees due to the used function approximation.

For each of the settings, we provide an experimental evaluation on diverse datasets and show that the method achieves a state-of-the-art performance. Also, it is flexible, robust, easy to use and can be improved with any domain independent advance in RL itself.

The article is structured as follows. First, in Section\;\ref{sec:problem}, we present the main ideas for solving various definitions of the problem, along with the problem of missing features. We explain the implementation details in Section\;\ref{sec:method}. In Section\;\ref{sec:experiments}, we describe the performed experiments and their results. Section\;\ref{sec:related_work} summarizes the related work.

\section{Problem variations} \label{sec:problem}
Before we delve into technical details, we present an overview of what CwCF is and how we view it. Then we start with the common notation which will be used for the rest of the article. In separate sections, we present the algorithms for the different cases. We start with the definition \eqref{eq:problem_original}, an average budget case where the budget is specified indirectly through a parameter $\lambda$. Next, still working with the average budget, we present a reformulation of the problem with a directly specified budget $b$ and solving it with the Lagrangian framework. Then we modify the framework to work with hard budgets. Lastly, we focus on a problem of missing features, which appear in many real-world situations.

\subsection{Classification with Costly Features}
\begin{figure}[t]
  \centering
  \includegraphics[width=1.0\linewidth]{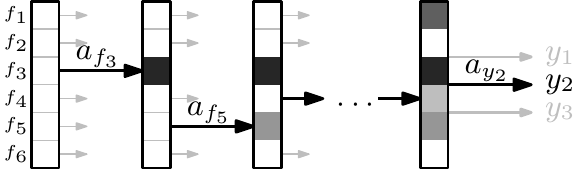}

  \caption{Illustration of the sequential process with a sample with 6 features ($f_1, ..., f_6$) and three classes ($y_1, y_2, y_3$). Feature values are acquired sequentially (actions $a_{f_3}$, $a_{f_5}$, ...) before making a classification ($a_{y_2}$). The particular decisions are influenced by the observed values -- the model chooses different actions for different samples.} 

  \label{fig:seq_process}
\end{figure}

First, we'd like to stress out the \emph{sequential} nature of the problem. Each sample is treated separately and the model sequentially selects features, one by one (see \fig{seq_process}). Eventually, a decision to classify is made, and the model outputs a class prediction. Each decision is based on the knowledge acquired so far, hence different samples will result in completely different sequences of features and predictions. This important fact differentiates CwCF from feature selection methods, where the same subset of features is selected for each sample. 

In real-world scenarios, there are many small modifications to the problem formulations. However, the presented method is very flexible and can be easily modified. For example, the prior knowledge can be included in the sample before starting the process (e.g., when a patient comes with known medical history). Multiple features can be grouped together and represented as one macro-feature. Different misclassifications can be treated with different weights through a particular choice of the loss function $\ell$. The method can also be used jointly with independently pretrained classifier and efficiently use it in situations where it works better (shown by \citet{janisch2019classification}).

\subsection{Common notation}
We assume that a sample can be represented as a real-valued vector, where each of its members we call \emph{features}. Here we assume a feature is one real number, but presented algorithms can be trivially modified in the case of multi-dimensional features.

Let's start with common notation, which will be used for the rest of the article. Let $(x,y) \in \mathcal{D}$ be a sample drawn from a data distribution $\mathcal{D}$. Vector $x \in \mathcal{X} \subseteq \mathbf{R}^n$  contains feature values, where $x_i$ is a value of feature $f_i \in \mathcal{F} = \{f_1, ..., f_n\}$, $n$ is the number of features, and $y \in \mathcal{Y}$ is a class. Let $c : \mathcal{F} \rightarrow \mathbf{R}^{+}$ be a function mapping a feature $f$ into its real-valued cost $c(f)$. For convenience, let's overload $c$ to also accept a set of features and return the summation of their individual costs: $c(\mathcal F') = \sum_{f \in \mathcal F'} c(f)$. Let $b \in \mathbf{R}^{+}$ be the allocated budget per sample. 

Our method selects features \emph{sequentially}, and is composed of a neural network with parameters $\theta$. However, for convenience, we define a pair of functions $(y_\theta, z_\theta)$ to represent the whole \emph{process} of classifying one sample. In this notation, $y_\theta: \mathcal{X} \rightarrow \mathcal{Y}$ represents the classification output at the end of the process and $z_\theta: \mathcal{X} \rightarrow \mathbf{R}$ represents the total cost of all features acquired during the process.

\subsection{Average budget with trade-off parameter $\lambda$}
\begin{figure}[t]
    \centering
    \includegraphics[width=0.6\linewidth]{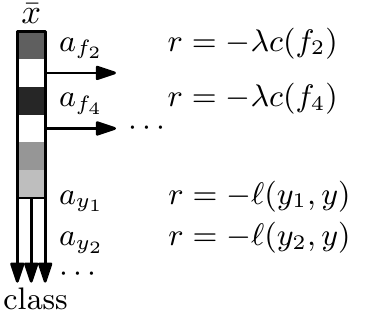}

    \caption{The MDP. The agent sees a masked sample $\bar x$. At each step it chooses from feature-selecting actions ($a_f$) or classification actions ($a_y$) and receives a corresponding reward (either the cost of the selected feature or the classification loss).}
    \label{fig:mdp}
\end{figure}

As we've seen in the medical example in the introduction, in some domains the user wants to target an average budget per sample. Let's start by writing the problem definition one more time:
\begin{equation*}
  \min_\theta \expect \big[ \ell(y_\theta(x), y) + \lambda z_\theta(x) \big]
  \tag{\ref{eq:problem_original} revisited}
\end{equation*}
Here, the user has to specify a trade-off parameter $\lambda$ which will result in an a priori unknown average budget. The approach is to create an MDP, where samples are classified in separate episodes and the expected reward $R$ per episode is:
$$  R = - \expect \big[ \ell(y_\theta(x), y) + \lambda z_\theta(x) \big] $$
Standard reinforcement learning techniques are then used to optimize this reward, thus solving \eqref{eq:problem_original}. Illustration of the MDP is in \fig{mdp}.

We model the environment as a deterministic MDP with full information, which is easily implemented. The agent, however, solves a stochastic MDP which is created when you remove some of the information (namely, the unobserved feature values). Formally, the MDP consists of states $\mathcal{S}$, actions $\mathcal{A}$, transition function $t$ and reward function $r$. State $s = (x, y, \mathcal{\bar{F}}) \in \mathcal{S}$ represents a sample $(x,y)$ and currently selected set of features $\mathcal{\bar{F}}$. The agent receives only the selected parts of $x$ without the label. Action $a \in \mathcal{A} = \mathcal{Y} \cup \mathcal{F}$ is either a classification action from $\mathcal{Y}$ that terminate the episode and the agent receives a reward of $-\ell(a, y)$, or a feature selecting action from $\mathcal{F}$ that reveals the corresponding value of $x$ and the agent receives a reward of $-\lambda c(a)$. The set of available feature selecting actions is limited to features not yet selected. Reward and transition functions are specified as:
$$ r(s, a) = 
          \begin{dcases*}
            -\lambda c(a)  & if $a \in \mathcal{A}_f$ \\
            -\ell(a, y)    & if $a \in \mathcal{A}_c$ 
          \end{dcases*}
          $$ 
$$
   t(s, a) = 
          \begin{dcases*}
            (x, y, \mathcal{\bar F} \cup a)   & if $a \in \mathcal{A}_f$ \\
            \mathcal{T}                       & if $a \in \mathcal{A}_c$
          \end{dcases*} $$

For real datasets, there may be a specific cost for missclassification, expressed in the amount of lost resources. If such information is not available, we propose to use a binary classification loss $\ell$:
$$ \ell(\hat y, y) = \begin{dcases*}
    0  & if $\hat y = y$ \\
    1  & if $\hat y \neq y$ 
  \end{dcases*}$$

\subsection{Average budget with specific target $b$}
As we already mentioned, a manual specification of an unintuitive parameter $\lambda$, as used in the previous section, isn't convenient. In real-world applications, the user wants to directly specify a budget $b$. Let's review the definition of the problem:
\begin{equation*}
  \min_\theta \expect \big[\ell(y_\theta, y) \big], \quad
  \text{s.t. } \expect \big[ z_\theta(x) \big] \leq b 
  \tag{\ref{eq:problem_average} revisited}
\end{equation*}
This constrained optimization problem can be transformed into an alternative Lagrangian form and solved with maxmin optimization. First, let's derive the Lagrangian, where $\lambda \in \mathbf{R}$ denotes a lagrange multiplier:
\begin{equation}\label{eq:lagrangian}
L(\theta, \lambda) = \expect \big[ \ell(y_\theta(x), y) + \lambda (z_\theta(x) - b) \big]
\end{equation}

The multiplier $\lambda$ plays a similar role as in the previous approach. However, here it is a variable of our algorithm and is \emph{not} specified by the user. As per \citet{bertsekas1999nonlinear}, there exist parameters $\theta, \lambda$ which are optimal in \eqref{eq:problem_average} and are a solution of the following problem:
\begin{equation}\label{eq:lagrange_solution}
  \max_{\lambda \geq 0} \min_\theta L(\theta, \lambda) 
\end{equation}
Inspired by an approach of \citet{chow2017risk}, we propose to iteratively perform gradient ascent in $\lambda$ and descend in $\theta$. For fixed $\theta$, optimizing $\lambda$ is easy, since the gradient is $\nabla_\lambda L = \expect \big[ z_\theta(x) - b \big]$. However, optimizing $\theta$ is not straightforward, since the model $(y_\theta, z_\theta)$ is neither differentiable nor continuous (it is a sequential process).
Let's look at the problem when $\lambda$ is fixed, that is, minimizing Lagrangian $L$ w.r.t. parameters $\theta$:
\begin{equation}
  \min_\theta L(\theta, \lambda) = \min_\theta \expect \big[ \ell(y_\theta(x), y) + \lambda z_\theta(x) \big] - \lambda b
\end{equation}
In the search for optimal parameters $\theta$, we can omit the term $\lambda b$ since it does not influence the solution. Note that the problem is then equal to \eqref{eq:problem_original} and thus we can directly apply RL through the method with fixed $\lambda$. However, we will only take small steps in $\theta$, effectively estimating and following the gradient $\nabla_\theta L$. The summary can be seen in Algorithm\;\ref{alg:lagrange}.

\begin{algorithm}[t]
\caption{Training with target budget $b$}
\label{alg:lagrange}
\begin{algorithmic}
\State $\lambda \leftarrow 0$, randomly initialize $\theta$
\Loop
  \State Update $\lambda$ by taking a gradient step with $\nabla_\lambda L = \expect \big[ z_\theta(x) - b \big]$ (maximize $L$)
  \State Update $\theta$ using RL (minimize $L$; Algorithms \ref{alg:rl} and \ref{alg:env})
\EndLoop
\end{algorithmic}
\end{algorithm}

A similar approach was evaluated in work of \citet{chow2017risk}, where the authors used the Lagrangian framework together with policy gradients to solve a constrained problem and proved convergence. Note that for an optimal solution, a stochastic policy may be needed. Our method is based on Q-learning, which works only with deterministic policies and this can result in oscillations around the stable point. However, we can detect when this happens, use it as a terminating condition and simply select the best-performing model satisfying the constraints.

\subsection{Hard budget}
In some domains, the resources are strictly restricted by a budget $b$ per sample. The problem definition changes to:
\begin{equation*}
  \min_\theta \expect \big[\ell(y_\theta, y) \big], \quad
  \text{s.t. } z_\theta(x) \leq b, \forall x 
  \tag{\ref{eq:problem_hard} revisited}
\end{equation*}
Similarly to the previous case, we can construct an MDP where the expected reward per sample is $R = -\expect\big[\ell(y_\theta(x), y)\big]$ and the episodes are restricted to end when the budget is depleted. Again, by solving this MDP with standard reinforcement learning techniques, we retrieve the solution to \eqref{eq:problem_hard}.

First, we change the reward function such that the costs of different features are ignored:
$$ r(s, a) = 
          \begin{dcases*}
            0                & if $a \in \mathcal{A}_f$ \\
            -\ell(a, y)      & if $a \in \mathcal{A}_c$
          \end{dcases*}
$$
Second, we restrict the set of available feature-selecting actions at each step to those, which do not exceed the specified budget. That is, $a \in \mathcal F$ is available only if $c(\mathcal{\bar F} \cup a) \leq b$. This way, the environment itself enforces the constraint. 

\subsection{Missing features} \label{sec:method_missing}
In a lot of domains, there is a large amount of data that can be used to train our method. However, the data is often not complete. I.e., in the medical domain, patients are typically sent only to a few examinations before the diagnosis is made. When using past data, only this limited information will be present in the training set. 

Here we present a principled method to deal with the issue, again by modifying our original algorithm. During training, a feature-selecting action is available only if the corresponding feature is present and the updates (see eq. \ref{eq:loss_q}) are made only with the estimates of available actions. We experimented with another variation, where estimates of all actions (even for unavailable features) were used. Intuitively, it corresponds to a case where we train with sparse data, but at test time, we have a full set. In our experiments, this approach underperformed the first one, hence we don't report it.

\section{Method} \label{sec:method}
In this section, we describe mainly the implementation of the reinforcement learning algorithm. Because we operate with large datasets with continuous features, the tabular approach is not feasible. Therefore, we employ neural networks as function approximators and use recent RL techniques. We experimented with a variety of different methods and found that incorporating recent insights from deep RL community is essential for the method to be stable, robust and perform well. After evaluating the implementation complexity and reported performance, we implemented Double Dueling DQN with Retrace as the RL solver. In the first part, we describe these RL methods. In the following part, we focus on the algorithm itself, and how it was implemented.

\subsection{Deep RL background}
An MDP is a tuple $(\mathcal{S}, \mathcal{A}, t, r, \gamma)$, where $\mathcal{S}$ represent state space, $\mathcal{A}$ a set of actions, $t(s, a)$ is a transition function returning a distribution of states after taking action $a$ in state $s$, $r(s, a, s')$ is a reward function and $\gamma$ is a discount factor. In Q-learning, one seeks the optimal function $Q^*$, representing the expected total discounted reward for taking an action $a$ in a state $s$ and then following the optimal policy. It satisfies the Bellman equation:
\begin{equation} \label{eq:beleq}
  Q^*(s, a) = \expect_{s' \sim t(s,a)} \left[r(s,a,s') + \gamma \max_{a'} Q^*(s', a') \right]
\end{equation}
A neural network with parameters $\theta$ takes a state $s$ and outputs an estimate $Q^\theta(s,a)$, jointly for all actions $a$. It is optimized by minimizing MSE between the both sides of eq. \eqref{eq:beleq} for transitions $(s_t, a_t, r_t, s_{t+1})$ empirically experienced by an agent following a greedy policy $\pi_\theta(s) = \argmax_a Q^\theta(s, a)$. Formally, we are looking for parameters $\theta$ by iteratively minimizing the loss function $\ell_\theta$, for a batch of transitions $\mathcal{B}$:
\begin{equation} \label{eq:loss}
\ell_\theta(\mathcal{B}) =  \expect_{(s_t,a_t,r_t,s_{t+1}) \in \mathcal{B}} \left[ q_t - Q^\theta(s_t, a_t) \right]^2
\end{equation}
where $q_t$ is regarded as a constant when differentiated, and is computed as:
\begin{equation} \label{eq:loss_q}
q_t = 
    \begin{dcases*}
      r_t                                             & if $s_{t+1} = \mathcal T$ \\
      r_t + \max_{a} \gamma Q^\theta(s_{t+1}, a)    & otherwise
    \end{dcases*}
\end{equation}
As the error decreases, the approximated function $Q^\theta$ converges to $Q^*$. However, this method proved to be unstable in practice \citep{mnih2015human}. Now, we briefly describe the techniques used in this work that stabilize and speed-up the learning.

\textbf{Deep Q-learning} \citep{mnih2015human} includes a separate \textit{target network} with parameters $\phi$, which follow parameters $\theta$ with a delay. Here we use the method of \citet{lillicrap2015continuous}, where the weights are regularly updated with expression $\phi := (1-\rho) \phi + \rho \theta$, with some parameter $\rho$. The slowly changing estimate $Q^\phi$ is then used in $q_t$, when $s_{t+1} \neq \mathcal T$:
\begin{equation} \label{eq:loss_dqn}
q_t = r_t + \max_{a} \gamma Q^\phi(s_{t+1}, a)
\end{equation}

\textbf{Double Q-learning} \citep{van2016deep} is a technique to reduce bias induced by the \textit{max} in eq. \eqref{eq:loss_q}, by combining the two estimates $Q^\theta$ and $Q^\phi$ into a new formula for $q_t$, when $s_{t+1} \neq \mathcal T$:
\begin{equation}
q_t = r_t + \gamma Q^\phi(s_{t+1}, \argmax_{a} Q^\theta(s_{t+1}, a))
\end{equation}
In the expression, the maximizing action is taken from $Q^\theta$, but its value is estimated with the target network $Q^\phi$.

\textbf{Dueling Architecture} \citep{wang2016dueling} uses a decomposition of the Q-function into two separate value and advantage functions. The architecture of the network is altered so that it outputs two estimates $V^\theta(s)$ and $A^\theta(s, a)$ for all actions $a$, which are then combined to a final output $Q^\theta(s, a) = V^\theta(s) + A^\theta(s, a) - \frac{1}{|\mathcal A|}\sum_{a'} A^\theta(s, a')$. When training, we take the gradient w.r.t. the final estimate $Q^\theta$. By incorporating baseline $V^\theta$ across different states, this technique accelerates and stabilizes training.

\textbf{Retrace} \citep{munos2016safe} is a method to efficiently utilize long traces of experience with truncated importance sampling. We store generated trajectories into an experience replay buffer \citep{lin1993reinforcement} and utilize whole episode returns by recursively expanding eq. \eqref{eq:beleq}. The stored trajectories are off the current policy and a correction is needed. For a sequence $(s_0, a_0, r_0, \dots, s_n, a_n, r_n, \mathcal T)$, we implement Retrace together with Double Q-learning by replacing $q_t$ with
\begin{equation} \label{eq:loss_q_full}
\begin{split}
q_t = r_t + \gamma \expect_{a \sim \pi_\theta(s_t)} \left[ Q^\phi(s_{t+1}, a) \right] + \\ + \gamma \bar\rho_{t+1} \left[q_{t+1} - Q^\phi(s_{t+1}, a_{t+1}) \right]
\end{split}
\end{equation}
where we define $Q^\phi(\mathcal T, \cdot) = 0$ and $\bar\rho_t = \min(\frac{\pi(a_t|s_t)}{\mu(a_t|s_t)}, 1)$ is a truncated importance sampling between exploration policy $\mu$ that was used when the trajectory was sampled and the current policy $\pi$. The truncation is used to bind the variance of product of multiple important sampling ratios for long traces. 
We allow the policy $\pi_\theta$ to be stochastic -- at the beginning, it starts close to the sampling policy $\mu$ but becomes increasingly greedy as the training progresses. It prevents premature truncation in the eq. \eqref{eq:loss_q_full} and we observed faster convergence. Note that all $q_t$ values for a whole episode can be calculated in $\mathcal O(n)$ time. Further, it can be easily parallelized across all episodes.

\subsection{Training method}
In this section, we describe the method of training the RL agent. At every step, the agent receives only an observation $o = \{ (x_i,f_i) \mid \forall f_i \in \mathcal{\bar{F}} \}$, that is, the selected parts of $x$ without the label.  The observation $o$ is mapped into a tuple $(\bar x, m)$:
$$
\bar{x}_i =
  \begin{dcases*}
     x_i   &  if $f_i \in \mathcal{\bar{F}}$ \\
     0     &  otherwise
  \end{dcases*}\quad ; \quad
m_i =
  \begin{dcases*}
     1    &  if $f_i \in \mathcal{\bar{F}}$ \\
     0    &  otherwise
  \end{dcases*}
$$
Vector $\bar{x} \in \mathbf{R}^n$ is a masked vector of the original $x$. It contains values of $x$ which have been acquired and zeros for unknown values. 
Mask $m \in \{0, 1\}^n$ is a vector denoting whether a specific feature has been acquired, and it contains 1 at a position of acquired features, or 0. The combination of $\bar{x}$ and $m$ is required so that the model can differentiate between a feature not present and observed value of zero. Each dataset is normalized with its mean and standard deviation and because we replace unobserved values with zero, this corresponds to the mean-imputation of missing values.

\begin{figure}[t]
  \centering
  \includegraphics[width=0.67\linewidth]{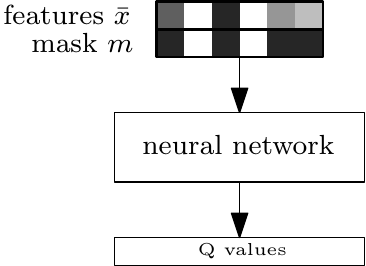}
  \caption{The architecture of the model. The input layer consists of the feature vector $\bar x$ concatenated with the binary mask $m$, followed by a feed-forward neural network (FFNN). Final fully connected layer jointly outputs Q-values for both classification and feature-selecting actions.}
  \label{fig:nn_arch}
\end{figure}

In our experiments, we use a feed-forward neural network, which accepts concatenated vectors $\bar{x}$, $m$ and outputs Q-values jointly for all actions. There are three fully connected hidden layers, each followed by the ReLu non-linearity, where the number of neurons in individual layers change depending on the used dataset. The overview is shown in Figure\;\ref{fig:nn_arch}.

A set of environments with samples randomly drawn from the dataset are simulated and the experienced trajectories are recorded into the experience replay buffer. After each action, a batch of transitions $\mathcal{B}$ is taken from the buffer and optimized upon with Adam \citep{kingma2014adam}, with eqs. (\ref{eq:loss}, \ref{eq:loss_dqn}). The gradient is normalized before back-propagation if its norm exceeds $1.0$. The target network is updated after each step. Overview of the algorithm and the environment simulation is in Algorithm \ref{alg:rl} and \;\ref{alg:env}.

\begin{algorithm}[t]
\caption{RL training}
\label{alg:rl}
\begin{algorithmic}
\State Randomly initialize parameters $\theta$
\State Pretrain the classifier part $Q^\theta(s,a \in \mathcal Y)$ with random states
\State Initialize target network $\phi \leftarrow \theta$
\State Initialize environments $\mathcal{E}$ with $(x, y, \emptyset) \in (\mathcal{X}, \mathcal{Y}, \wp(\mathcal{F}))$
\State Initialize replay buffer $\mathcal{M}$ with a random agent
\Loop
  \ForAll {$e \in \mathcal{E}$}
    \State Simulate one step with $\epsilon$-greedy policy $\pi_\theta$:
      $$a = \pi_\theta(s);\quad s', r = \Call{step}{e, a}$$
    \State Add transition $(s, a, r, s')$ into circular buffer $\mathcal{M}$
  \EndFor
 
  \State Sample a random batch $\mathcal{B}$ from $\mathcal{M}$
  
  \ForAll {$(s_i, a_i, r_i, s_i') \in \mathcal{B}$}
    \State Compute target $q_i$ according to eq. \eqref{eq:loss_q_full}
    \State Clip $q_i$ with maximum of $0$
  \EndFor

  \State Perform one step of gradient descent on $\ell_\theta$ w.r.t. $\theta$:
  $$\ell_\theta(\mathcal{B}) = \sum_{i=1}^{|\mathcal{B}|} (q_i - Q^\theta(s_i, a_i))^2$$

  \State Update parameters $\phi := (1-\rho) \phi + \rho \theta$
\EndLoop
\end{algorithmic}
\end{algorithm}

\begin{algorithm}[t]
\caption{Environment simulation}
\label{alg:env}
\begin{algorithmic}
\State {\footnotesize Operator $\odot$ marks the element-wise multiplication.}
\Function{Step}{$e \in \mathcal{E}$, $a \in \mathcal{A}$}
    \If{$a \in \mathcal{A}_c$}
      \State $ r = 
        \begin{dcases*}
           0 &  if $a = e.y$ \\
          -1 &  if $a \neq e.y$
        \end{dcases*}$
      \State Reset $e$ with a new sample $(x, y, \emptyset)$ from a dataset
      \State Return $(\mathcal{T}, r)$
    \ElsIf{$a \in \mathcal{A}_f$}
      \State Add $a$ to set of selected features: $e.\mathcal{\bar{F}} = e.\mathcal{\bar{F}} \cup a$
      \State Create mask $m_i=1$ if $f_i \in \mathcal{\bar{F}}$ and $0$ otherwise
      \State Return $((e.x \odot m, m), -\lambda c(a))$
    \EndIf
\EndFunction
\end{algorithmic}
\end{algorithm}

Because all rewards are non-positive, the whole Q-function is also non-positive. We use this knowledge and clip the $q_t$ value so that it is at most $0$. Without this bound, the predicted values sometimes rose to infinity, due to the \emph{max} used in Q-learning. The definition of the reward function also results in optimistic initialization. A neural network with initial weights tends to output small values around zero. Effectively, the model tends to overestimate the real Q-values, which has a positive effect on exploration.

We don't use a discount factor ($\gamma = 1$), because we want to recover the original objectives. We use $\epsilon$-greedy policy that behaves greedily most of the time, but picks a random action with a probability $\epsilon$. Exploration rate $\epsilon$ starts at a defined initial value and it is linearly decreased over time to its minimum value. 

Classification actions $\mathcal A_c$ are terminal and their Q-values do not depend on any following states. Prior to the main method, we pretrain the part of the network $Q^\theta(s, a)$, for classification actions $a \in \mathcal A_c$ with batches of randomly sampled states. We randomly pick samples $x$ from the dataset and generate masks $m$. The values $m_i$ follow the Bernoulli distribution with probability $p$. As we want to generate states with different amount of observed features, we randomly select $\sqrt[3]{p} \sim \mathcal U(0, 1)$ for different states. The resulting distribution of states is shifted towards the initial state with no observed features. The main algorithm starts with accurate classification predictions and this technique has a positive effect on the speed of the training process.

In the case of the specified budget $b$, we also optimize the multiplier $\lambda$. In our experiments, we found a simple gradient ascent with momentum works best.
The learning rate schedule for both parameters $\theta$ and $\lambda$ is exponential, in fixed steps, up to some minimal value.

\section{Experiments} \label{sec:experiments}
In this section, we describe the performed evaluation of the methods described in Section\;\ref{sec:problem}. The code used in this evaluation will be available at \url{https://github.com/jaromiru/cwcf}.

\subsection{Evaluation metric} \label{sec:metric}
\begin{figure}[t]
    \centering
    \includegraphics[width=0.6\linewidth]{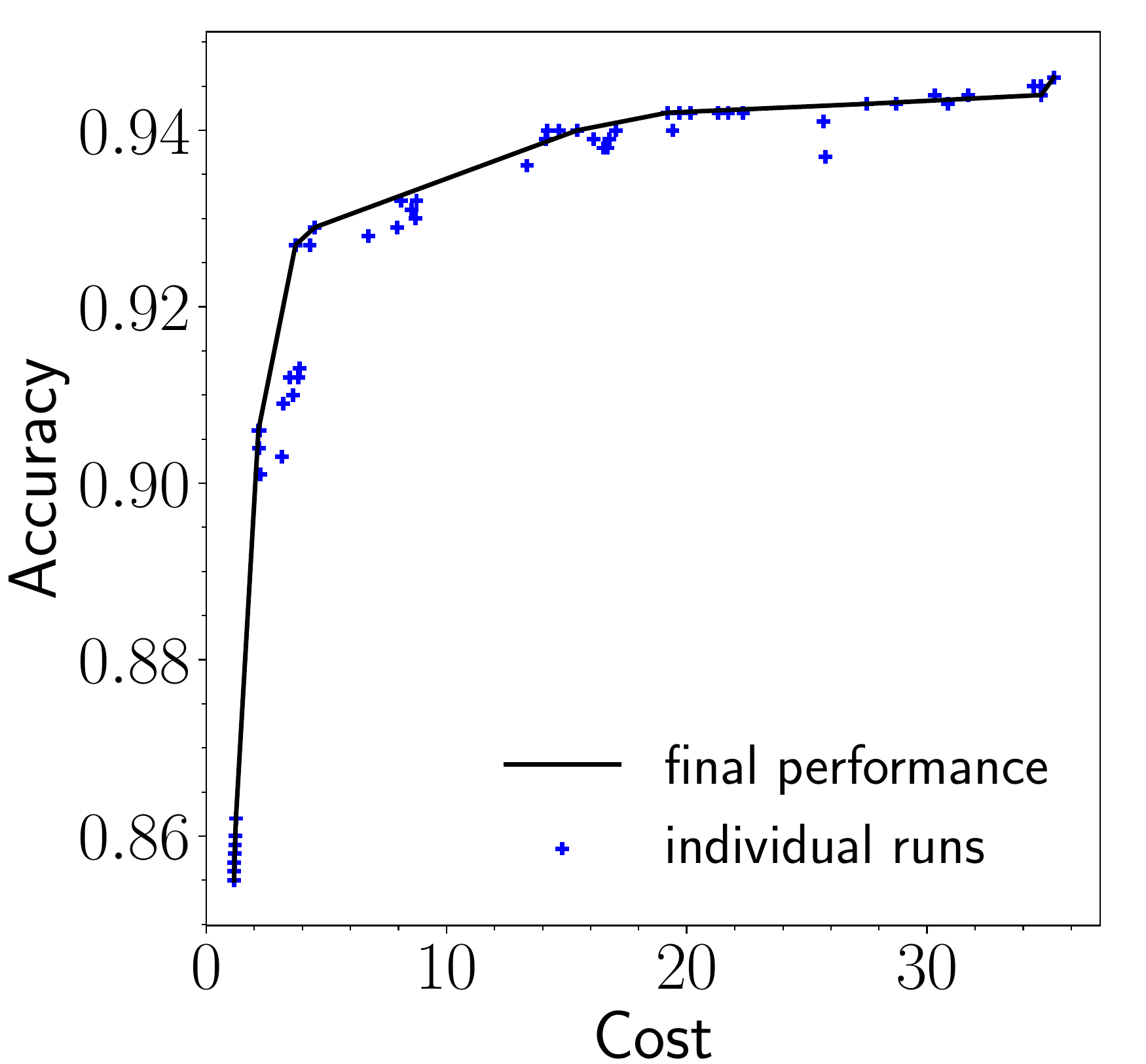}

    \caption{Illustrative performance of different trained models and their trade-offs, measured on the cost-accuracy plane. \emph{Validation set} is used to select the best performing models, hence the individual runs can sometimes exceed the final performance, which is reported on the test set.}
    \label{fig:mb_scatter}
\end{figure}

It's difficult to compare algorithms when we essentially optimize for two objectives - cost and accuracy. Thus, we adopt the following procedure. We train multiple instances of a particular algorithm, with varying parameters (this involves different settings of $\lambda$, budget $b$ and seeds). The exact number of instances differs across datasets, settings and algorithms, but is comparable, with median of 20. In the cost-accuracy plane, we use the \emph{validation set} to select the best performing model instances, which form a convex hull over all trained models. As an example, see \fig{mb_scatter}, where we show several trained models and the selected ones. Note that because we select the best points on the validation set, occasionally some points can be higher that the final curve. For the final metric, we use the normalized area under this curve. By normalization we mean division by the area of the whole cost-accuracy plane, such that the best value is 1.0 (higher is better). We assume that, for each dataset, all models can achieve prior accuracy with no features and also the maximal accuracy of a particular model with all features. 

\subsection{Baseline method}
We design a simple baseline method to compare with. First, we use a feature selection technique to select a \emph{fixed order} of features, sorted from most important to least. Then, we iteratively add features, according to the list, and train separate neural network based classifiers. The resulting performance is visualized at the cost-accuracy graph as usual. Note that this baseline can be compared both to average and hard budget methods since for every budget, the set of used features is fixed.
More specifically, we use Recursive Feature Elimination together \citep{guyon2002gene} with Ridge classifier \citep{hoerl1970ridge} to select the feature order. The size of the neural network is comparable to the neural network used in the main method for a particular dataset.

\subsection{Used datasets}
In the following sections, we use several datasets, information about which is summarized in Table\;\ref{tab:datasets}. They were obtained from public sources \citep{uciml,krizhevsky2009learning} and the diabetes dataset was obtained from the authors of prior work \citep{kachuee2019opportunistic}. For datasets where there are no explicit costs, we use uniform costs for all features. Miniboone dataset is small and easy, from the classification perspective, and it is suitable for fast experimenting and evaluation. Forest dataset contains categorical features and many samples, making it hard to achieve good performance. Cifar and mnist datasets are challenging multiclass image recognition datasets, where we treat all pixels as separate features. We could leverage convolutions for the image datasets, but to make a fair comparison with other algorithms, we treat all datasets the same -- as with features with no clear structure. Diabetes dataset contains real-world medical data with expert-valued feature costs and we use its balanced and mean imputed version.

\begin{table}[b]
    \centering
    \begin{tabular}{lrrrrrc}
      \toprule
      Dataset     & \#feats    & \#class & \#trn  & \#val     &  \#tst  & costs     \\
      \midrule                                
      miniboone   & 50        & 2         & 45k       & 19k       & 65k       & U         \\
      forest      & 54        & 7         & 200k      & 81k       & 300k      & U         \\
      cifar       & 400       & 10        & 40k       & 10k       & 10k       & U         \\
      mnist       & 784       & 10        & 50k       & 10k       & 10k       & U         \\
      diabetes     & 45        & 3         & 64k       & 14k       & 14k       & V         \\
      \bottomrule
    \end{tabular}

    \caption{Used datasets. The cost is either uniform (U) or variable (V) across features.}
    \label{tab:datasets}
\end{table}

\subsection{Experimental setup}
All evaluated algorithms include a $\lambda$ like trade-off parameter, which we sweep across different values and run the algorithms several times, with different seeds. We use the evaluation method described in Section\;\ref{sec:metric} to present the results. 

As for our algorithm, we let it run for a pre-defined number of steps, which is dependent on the particular dataset. For each dataset, we only define two hyperparameters, \emph{ep\_len} (as for epoch length) and the size of layers in the neural network. All other hyperparameters are either the same or derived from the \emph{ep\_len} parameter. We also have to highlight the fact that the hyperparameters stay the same across all versions of our algorithm, clearly featuring its robustness. Table \ref{tab:parameters} show all used parameters.

\subsection{Average budget with trade-off $\lambda$}
\citet{janisch2019classification} already established the result that RL based methods are superior to prior-art techniques. It performed robustly on all tested datasets, comparable or better than prior-art. Here we select four representative datasets and compare the RL method (\emph{rl}) in average budget settings with Adapt-Gbrt (\emph{agbrt}) \citep{nan2017adaptive} and Budget-Prune (\emph{bprune}) \citep{nan2016pruning}, where applicable.

\begin{figure}[t]
  \centering
  {
  \setlength{\tabcolsep}{0pt}
  \begin{tabular}{ccccc}
    \includegraphics[width=0.33\linewidth]{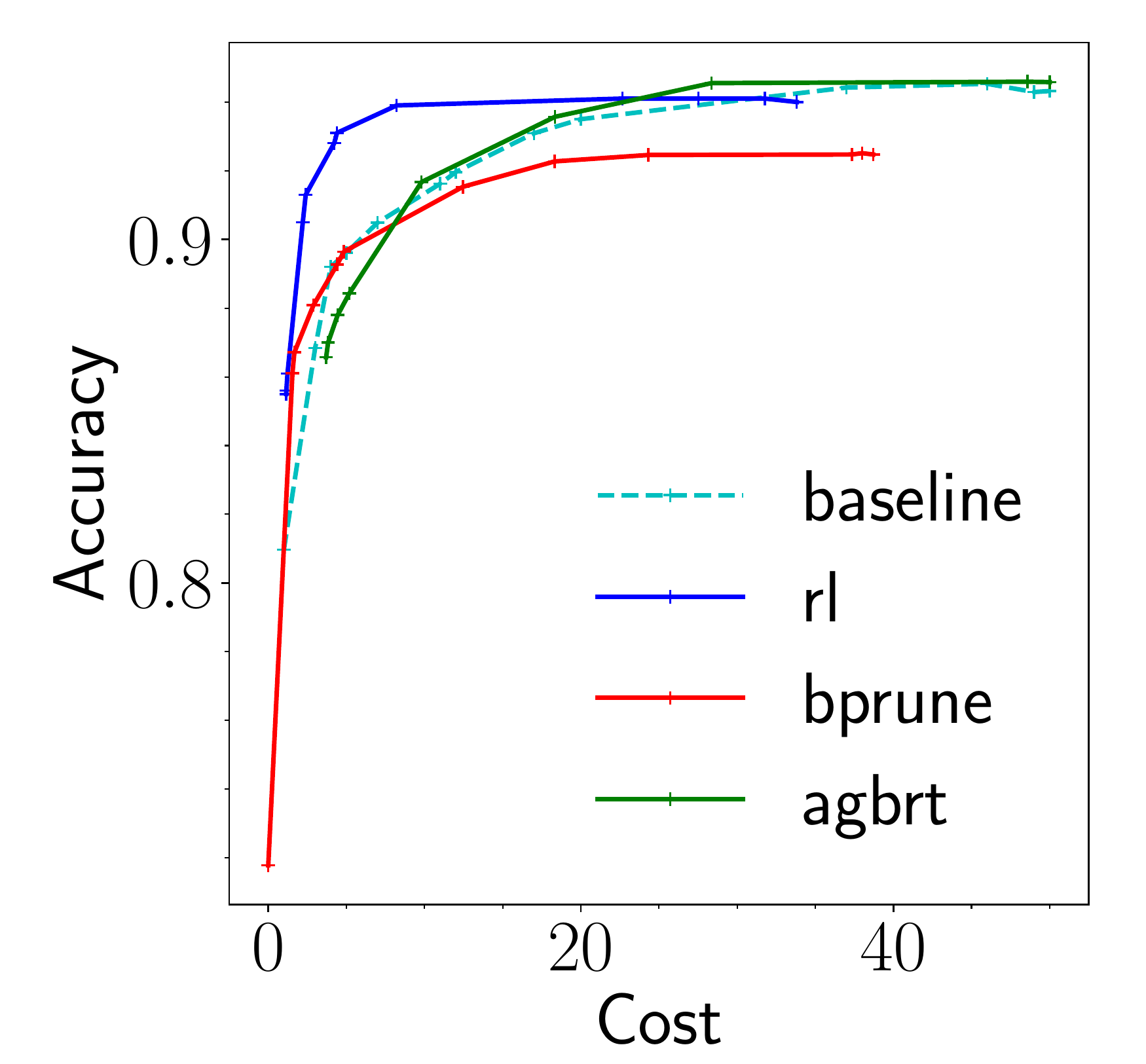} &
    \includegraphics[width=0.33\linewidth]{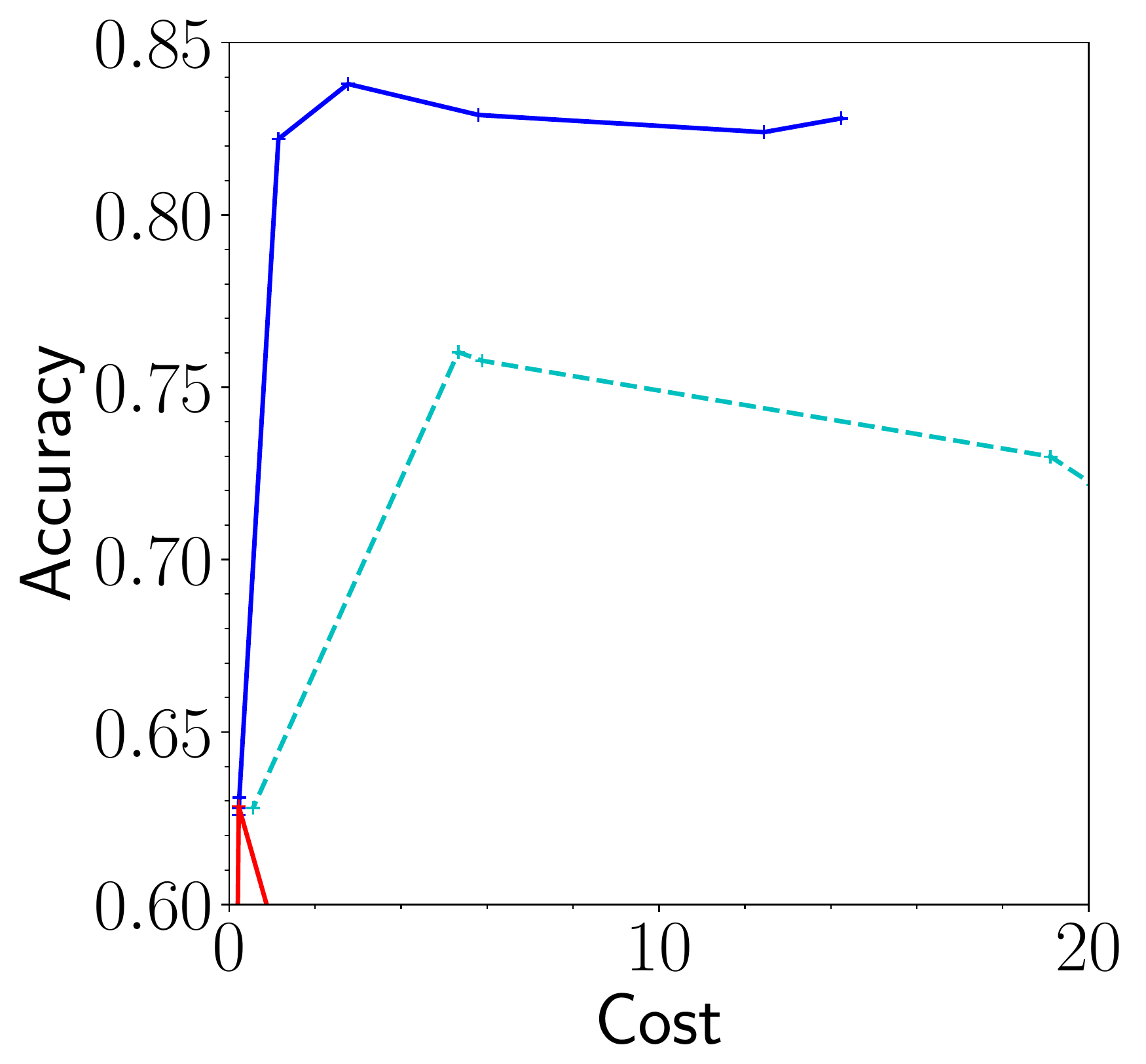} &
    \includegraphics[width=0.33\linewidth]{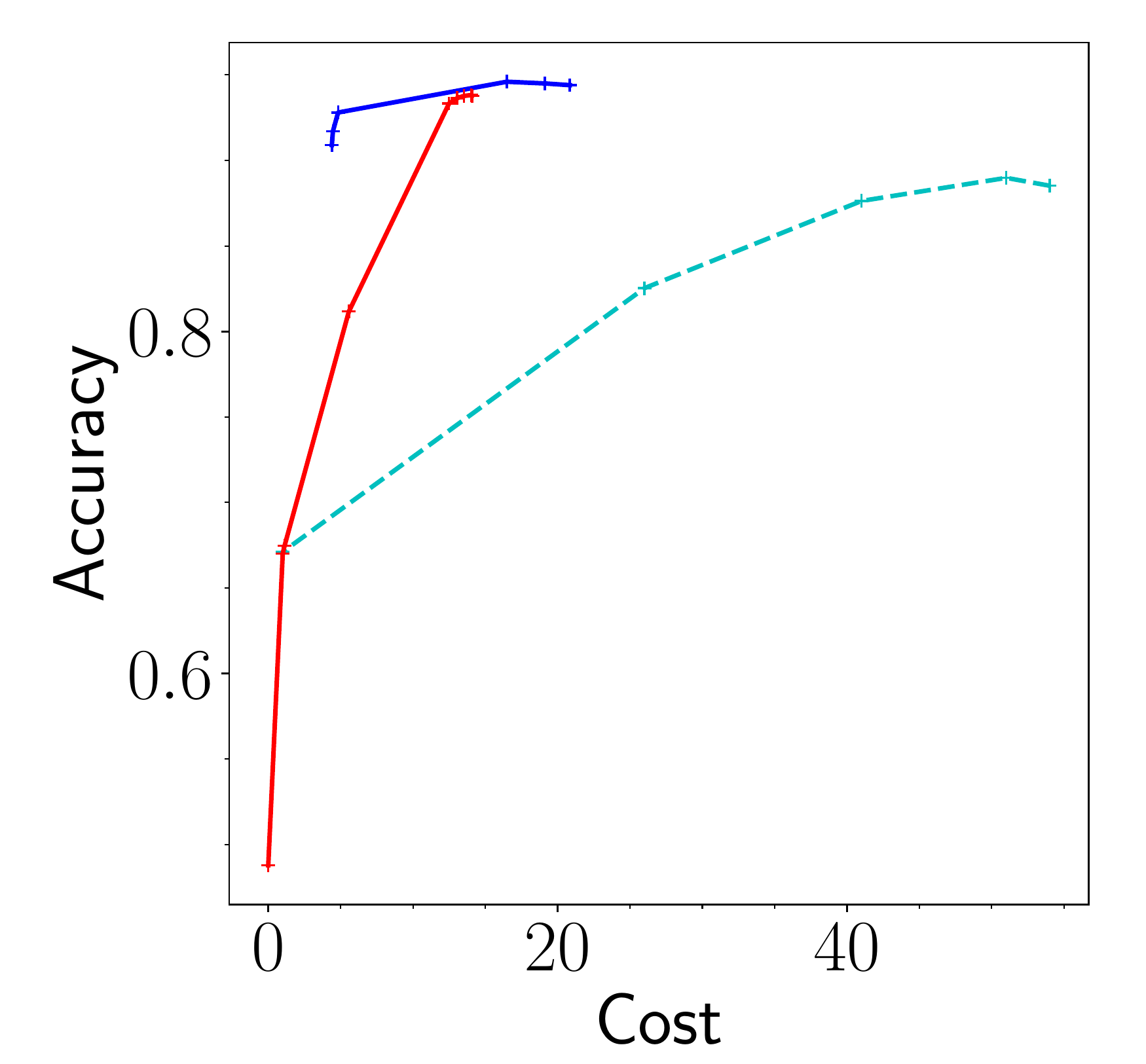} \\
    (a) miniboone & (b) diabetes & (c) forest \\
    
    \includegraphics[width=0.33\linewidth]{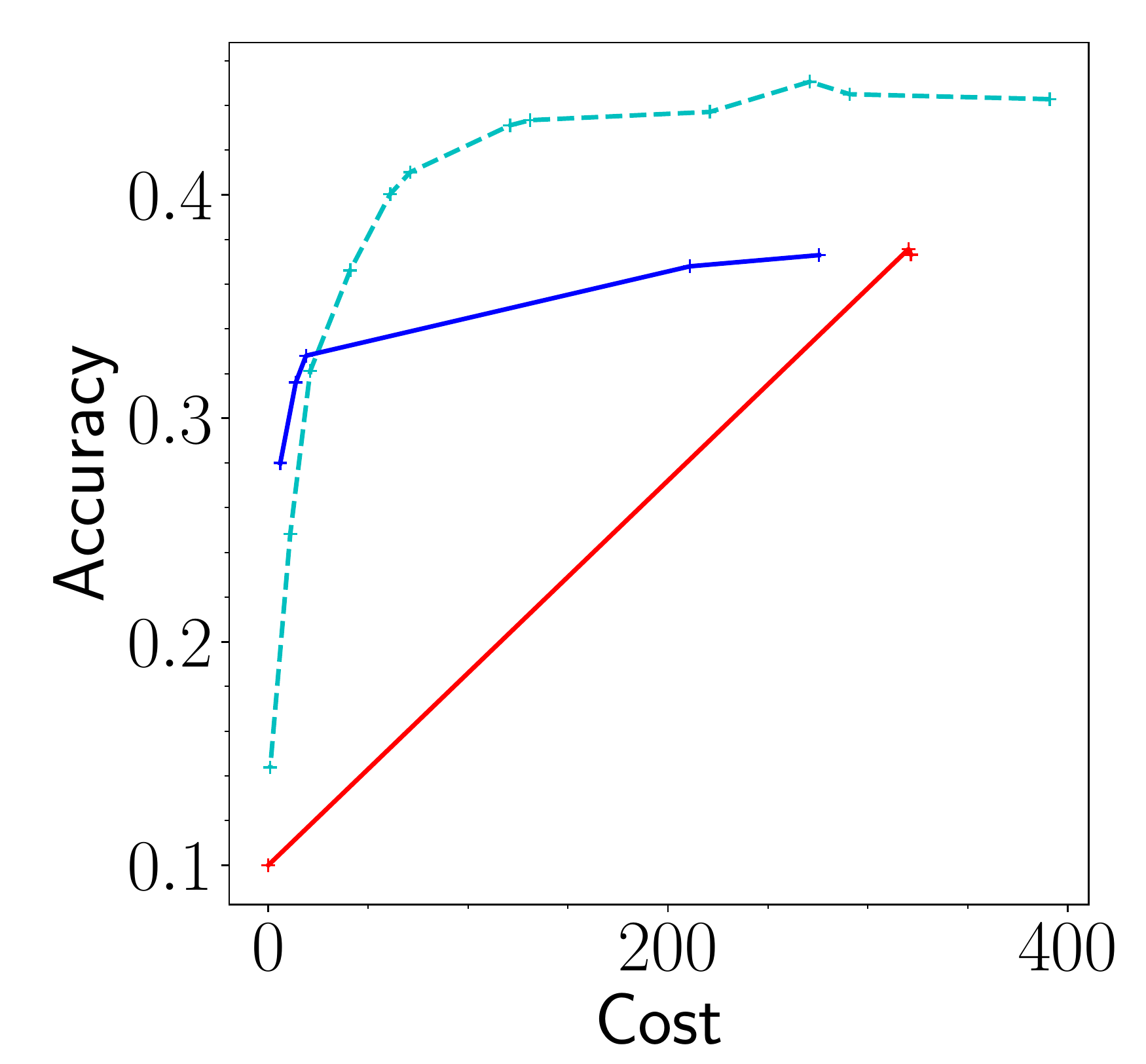} &
    \includegraphics[width=0.33\linewidth]{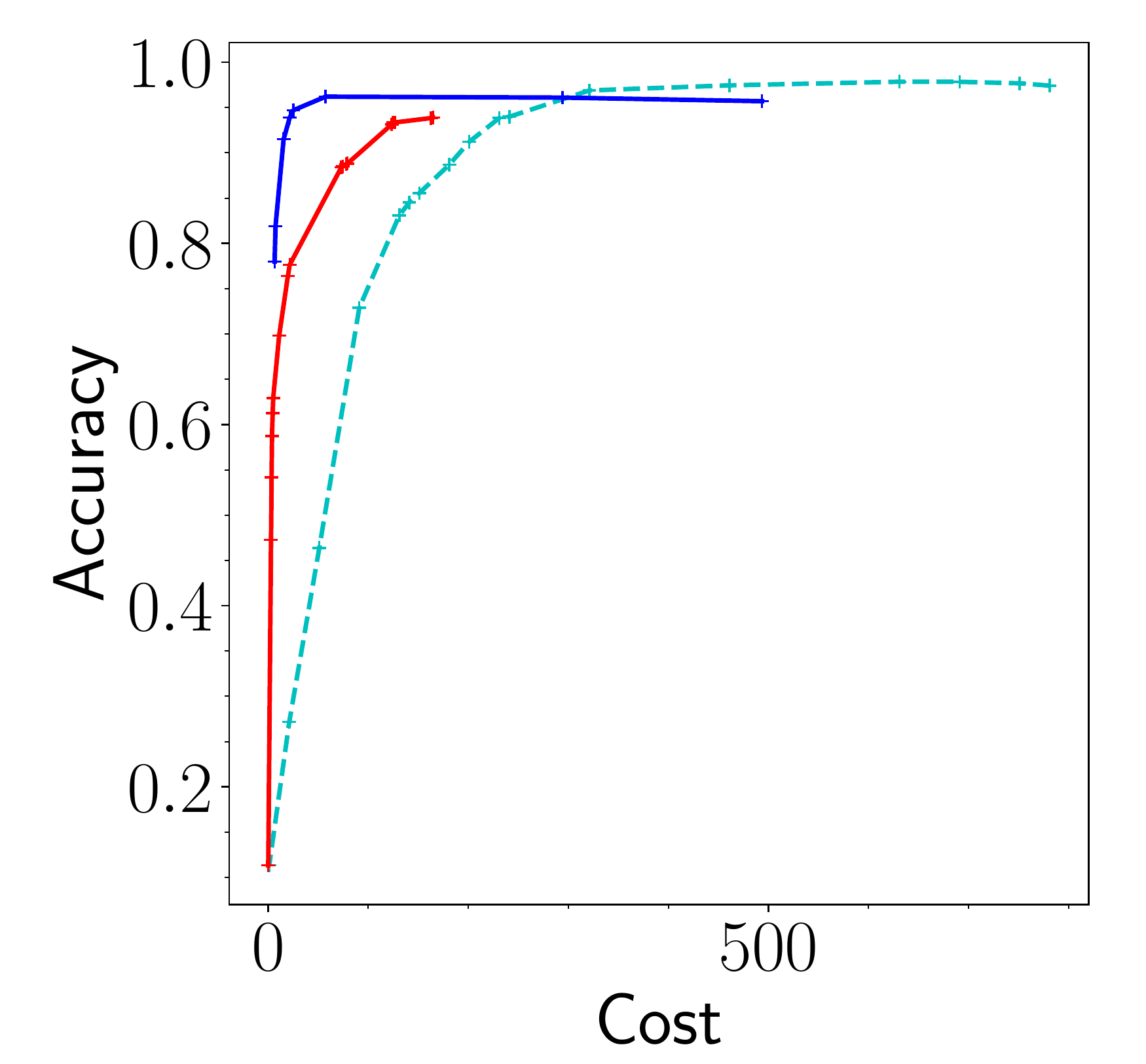} \\
    (d) cifar & (e) mnist \\
  \end{tabular}
  }
  \vspace*{2mm}

  \begin{tabular}{lrrrr}
    \toprule
                & baseline        & rl     & agbrt     & bprune    \\
    \midrule                                
    miniboone   & 0.925     & \textbf{0.935}     & 0.924     & 0.914   \\
    diabetes    & 0.748     & \textbf{0.829}     & N/A       & 0.628   \\
    forest      & 0.806     & \textbf{0.924}     & N/A       & 0.906   \\
    cifar       & \textbf{0.419}     & 0.354     & N/A       & 0.265   \\
    mnist       & 0.888     & \textbf{0.954}     & N/A       & 0.921   \\
    \bottomrule
  \end{tabular}

  \caption{Results in the average budget setting, trained through $\lambda$-specified budget. The table shows the normalized area under the trade-off curve as the overall metric. Adapt-Gbrt algorithm cannot be evaluated in multi-class datasets.}
  \label{fig:exp_avg}
\end{figure}

Adapt-Gbrt is a random forest (RF) based algorithm that uses an external pretrained model (HPC). It jointly learns a gating function and Low-Prediction Cost model (LPC) that adaptively approximates HPC in regions where it suffices for making accurate predictions. The gating function then redirects the samples to either use HPC or LPC. BudgetPrune is another algorithm that prunes an existing RF using linear programming to optimize for the cost vs. accuracy trade-off. We obtained the results for both Adapt-Gbrt and BudgetPrune by running the source code published by their authors. The published version of Adapt-Gbrt is restricted to datasets with only two classes.

The results are shown in \fig{exp_avg}. As already stated, RL provides superior performance in all datasets, compared to prior-art. Budget-Prune simply fails in case of the diabetes dataset, as it overfits the training set almost perfectly with a budget of about 5, but its test accuracy is only 0.42. In miniboone, it is noteworthy that the Adapt-Gbrt and Budget-Prune algorithms do not exceed the performance of the baseline classifier. In the cifar dataset, the baseline method provides well and consistent performance across all budgets, exceeding other methods by a large margin. It is only surpassed by RL when small budgets (up to 20 features) are targeted. We assume that the model capacity is the restricting factor here, as cifar is a very hard dataset, especially when pixel relations are disregarded. Note that the baseline classifier solves much easier task - it is trained with a static set of pixels for each budget. On the other hand, RL has to learn all possible permutations of available pixels, which is a much harder task. Note that we don't use convolutions, which are common in image recognition tasks (to regard all datasets the same), but they could be incorporated into the algorithm, if needed.

One thing that needs to be taken into account is the convergence speed of the RL method. The difficulty of the dataset, number of features, classes and samples influence the time needed. On a host equipped with Intel Xeon E5-2650v2 2.60GHz and nVidia Tesla K20 5GB GPU, it takes about 20 minutes to run the algorithm in the miniboone and diabetes datasets (for one specific $\lambda$). In the forest dataset, it is about 55 hours. Cifar needs about 5 days and mnist about 10 days. Evaluation of a trained model is fast and takes a negligible amount of time.

\subsection{Average budget with target $b$}
Performance-wise (using the defined metric), the results are similar to the $\lambda$ defined budget. The previous methods can be applied if we simply want to sweep across all spectrum of budgets, e.g., for comparison reasons. However, if the user targets a particular budget $b$, this method is highly preferred, because there is not additional parameter.

\begin{figure}[t]
    \centering
    \begin{tabular}{cc}
      \includegraphics[width=0.45\linewidth]{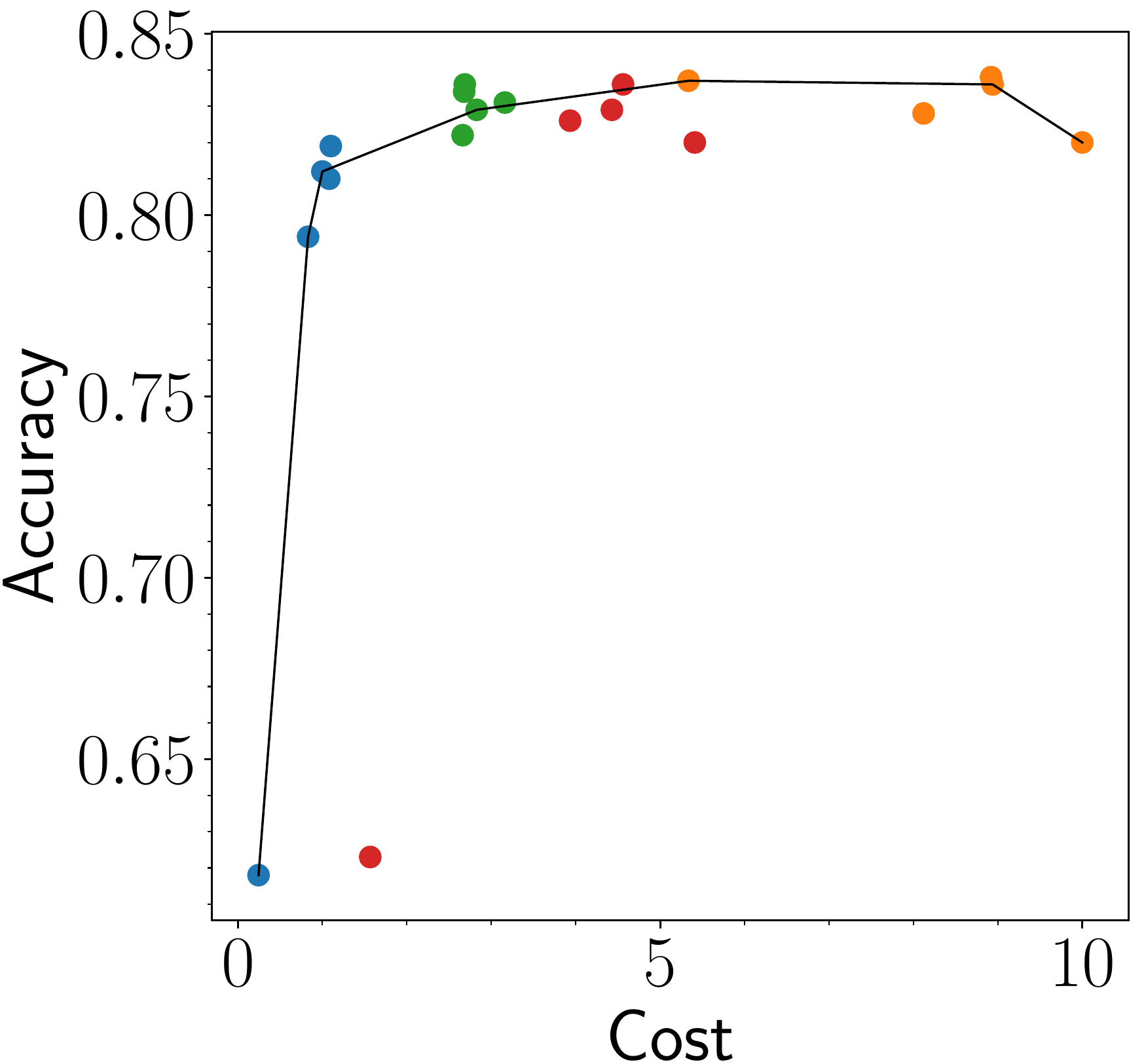} &
      \includegraphics[width=0.45\linewidth]{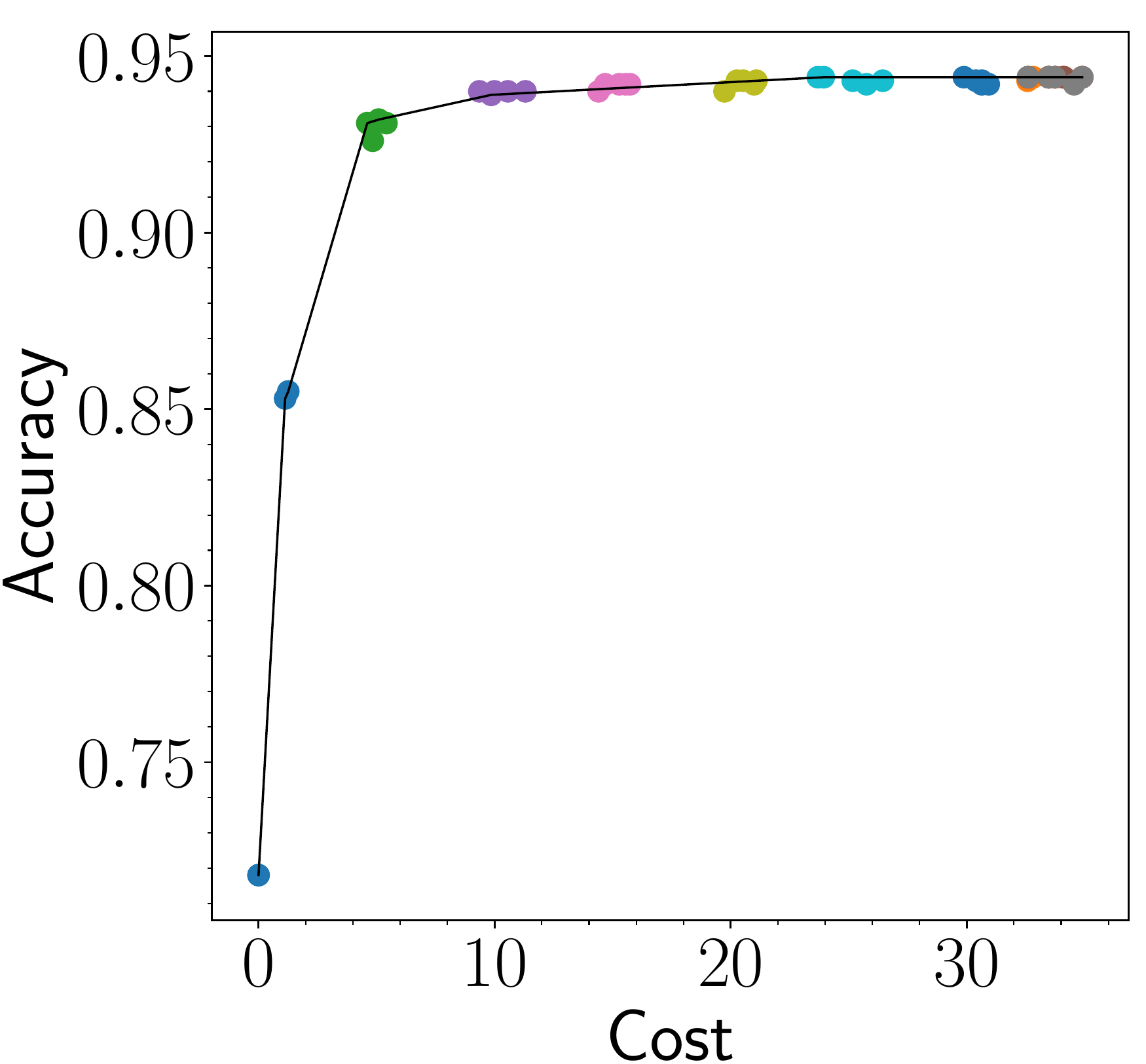} \\

      (a) diabetes & (b) miniboone
    \end{tabular}

    \caption{Average budget setting, with directly specified budget $b$. (a) Five runs with different seeds for each budget of $\{1, 3, 5, 10\}$ in the diabetes dataset. (b) Five runs for each budget $\{0, 5, ..., 45, 50\}$ in the miniboone dataset.
    The different budgets are plotted with different colors, showcasing the variance between runs.}
    \label{fig:mb_avg_budget}
\end{figure}

We selected two small datasets, miniboone and diabetes to conduct the following experiment. In miniboone, we evaluated 5 runs for each budget between 0 to 50, with a step of 5 and plotted them on the cost-accuracy plane with different colors, to highlight the variance between different runs. In diabetes, we selected the budgets as \{1, 3, 5, 10\}, because this is the most interesting part of the trade-off curve. In \fig{mb_avg_budget}, we can see the results.
In the case of the miniboone dataset, all runs with the same targeted budget resulted in very similar performance. In the diabetes dataset, the variance of the runs is comparatively higher. The experiment suggests that to obtain good results in practice, the method should be run several times and only the best performing model should be selected (based on the validation set).

We noted that the learned models always use the whole available budget up to some point, where it cannot strengthen its accuracy, even with more features. The observation is consistent with previous experiments with $\lambda$-targeted budget, where further lowering $\lambda$ did not improve accuracy nor depleted more budget. Miniboone dataset has 50 features, and as it can be seen in the \fig{mb_avg_budget}b, the model retrieved 35 at most. Cost of all features in the diabetes dataset is about 20.5, but the model stops acquiring at the cost of about 13.

\begin{figure}[t]
    \centering
    \includegraphics[width=1.0\linewidth]{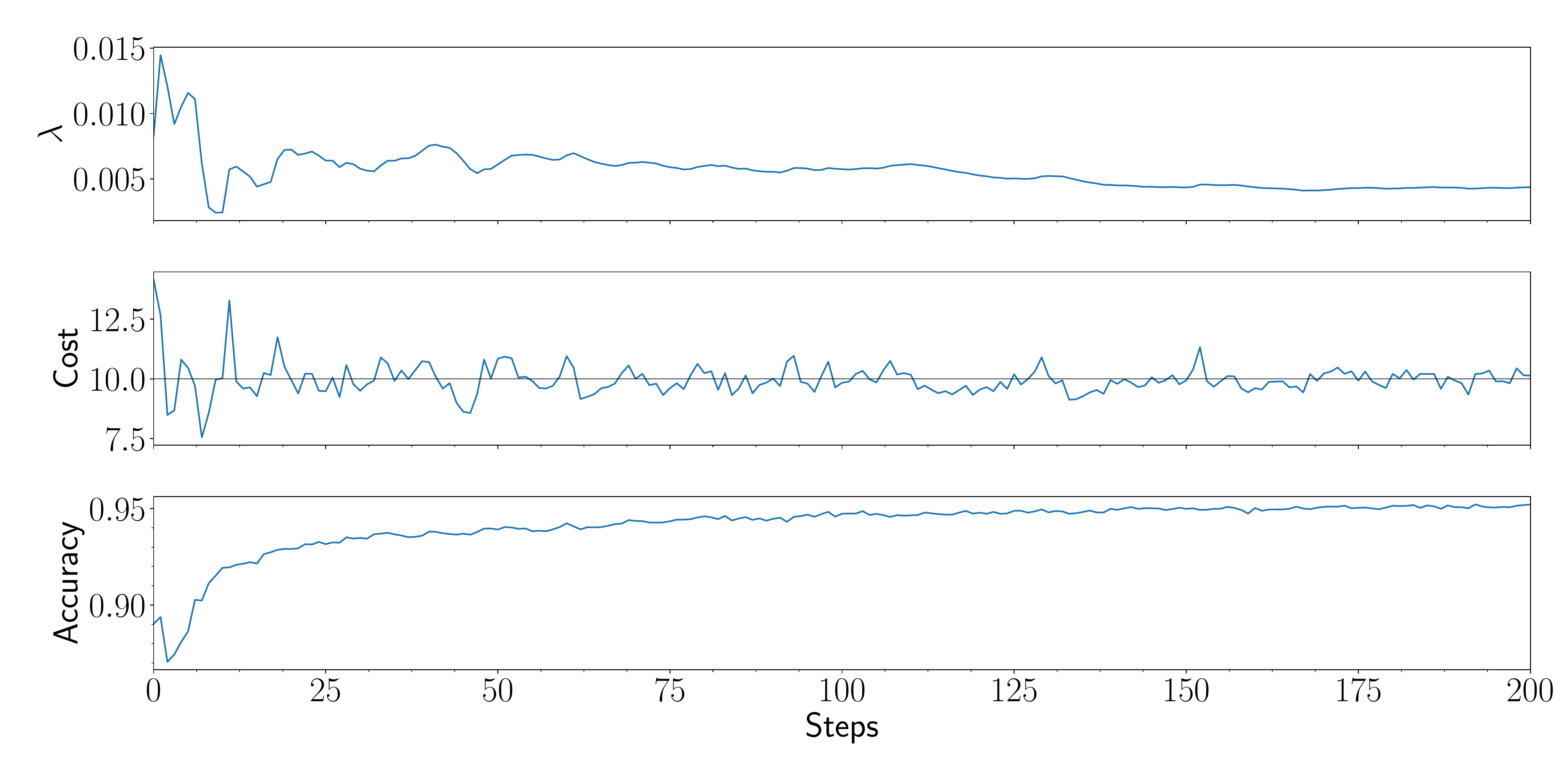}

    \caption{The learning progress in the miniboone dataset, with an average target budget $b=10$. The plot shows changes in $\lambda$, spent budget and accuracy during learning (one run, not averaged). One step on the $x$ axis corresponds to 100 training steps.}
    \label{fig:lambda_progress}
\end{figure}

We also analyze the training progress in \fig{lambda_progress}. At first, the $\lambda$ multiplier oscillates, until it converges to its optimal value. Similar oscillations can also be seen in the budgets spent by partially trained models, where the budget approaches the target value by the end of the training. We assume that a small deviation from the average target budget is acceptable. If not, we can simply select the last model that strictly meets the constraint.

\begin{figure}[t]
    \centering
    \includegraphics[width=1.0\linewidth]{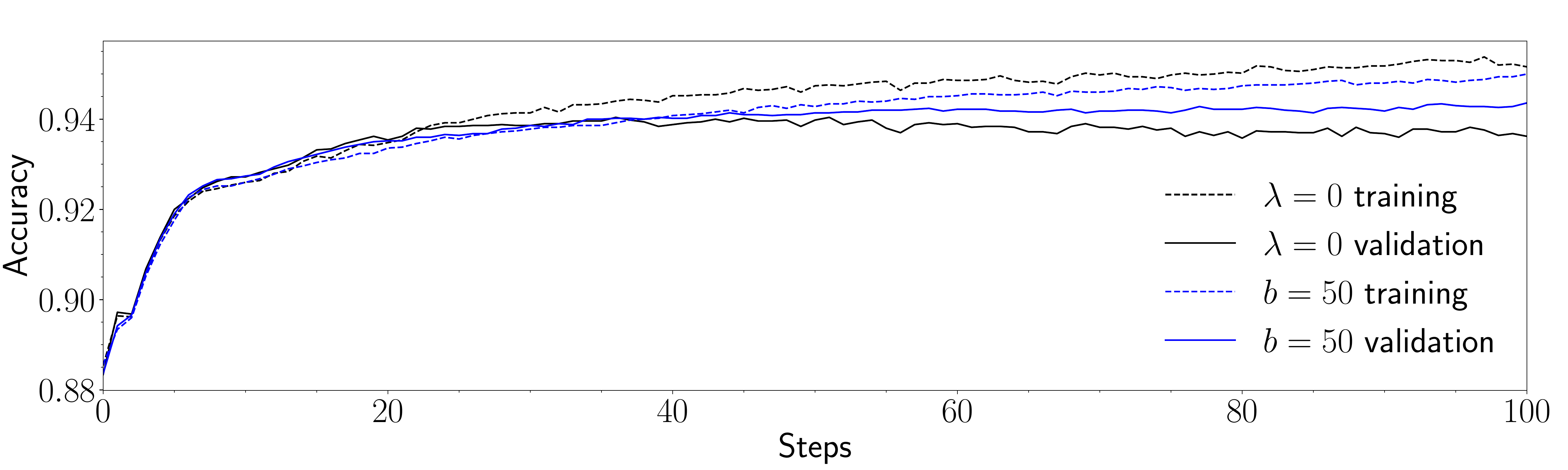}

    \caption{Comparison of $\lambda$-set budget and specific target $b$. Theoretically equal settings, $\lambda = 0$ (meaning free features) and a specific budget target $b = 50$ (all features in the dataset), result into different behavior. All other settings are the same. Averaged over five runs, one step on the $x$ axis corresponds to 100 training steps.}
    \label{fig:lmb_vs_budget}
\end{figure}

In \fig{lmb_vs_budget}, we analyze a training progress of the two methods, when a budget is specified directly and indirectly with $\lambda$. With miniboone dataset, we selected two, in theory, equal settings: $\lambda = 0$ and $b=50$. With fixed $\lambda$, setting it to zero effectively means that all features are free and budget is infinite. In the other case, setting $b=50$ means that all features can be acquired (there are 50 of them and the cost is uniformly 1.0). All other settings were equal and we conducted 5 runs of each algorithm and averaged their results. \fig{lmb_vs_budget} shows that the specific $b$-budget method is more resilient to over-fitting -- while its performance raises slowly than in $\lambda$-set budget, the validation performance monotonically raises as well. We speculate that the changes in $\lambda$ multiplier inhibits over-fitting by forcing the model to dynamically adapt to the non-stationary environment. Also, the asymptotic average accuracy is better in the case of $b$-budget, about 0.943 in 20000 steps while $\lambda$-budget reaches its top accuracy of 0.940 in about 3500 steps. As a corollary, with $\lambda$-budget a validation performance has to be tracked to prevent over-fitting, while with $b$-budget, the algorithm can be just run with a fixed number of steps.

In conclusion, using the specific budget $b$ method has several advantages. It achieves slightly higher accuracy, displays better over-fitting resiliency and avoids a superfluous hyperparameter.

\subsection{Hard budget}

\begin{figure}[t]
  \centering
  {
  \setlength{\tabcolsep}{0pt}
  \begin{tabular}{cccc}
    \includegraphics[width=0.45\linewidth]{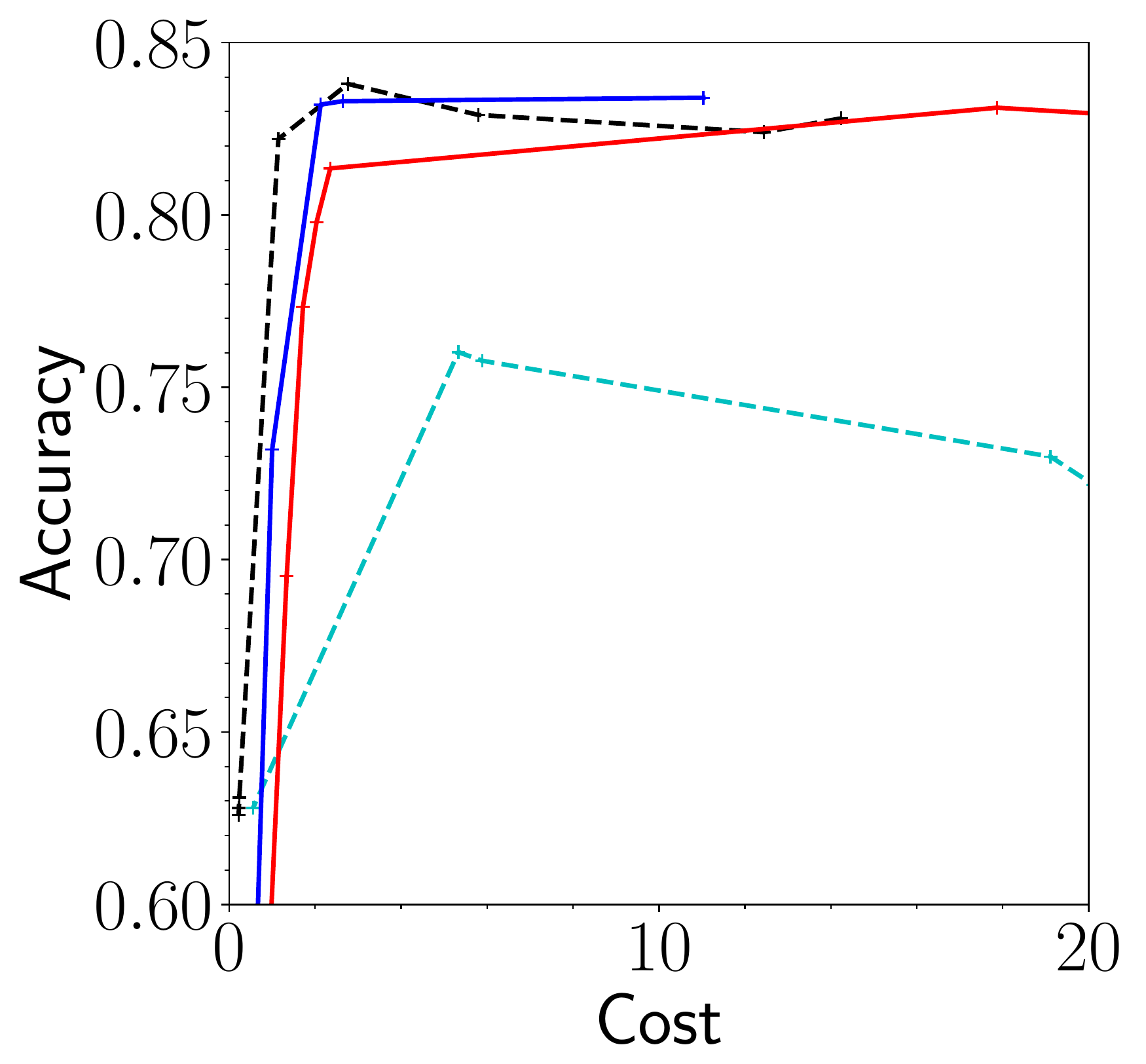} &
    \includegraphics[width=0.45\linewidth]{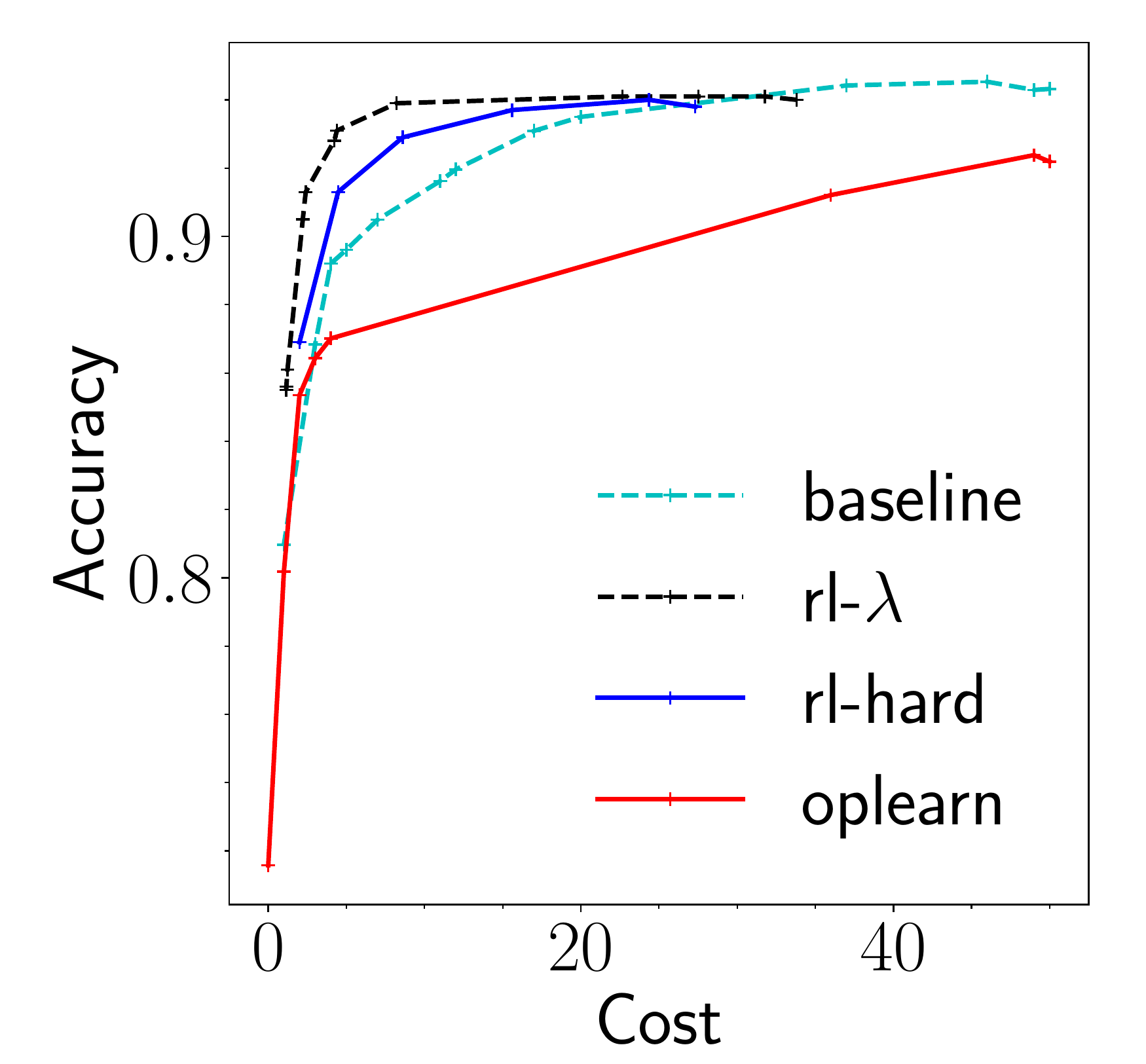} \\

    (a) diabetes & (b) miniboone \\
  \end{tabular}
  }
  \vspace*{2mm}

  \begin{tabular}{lrrr}
    \toprule
    Dataset     & baseline        & rl-hard      & oplearn    \\
    \midrule                                
    diabetes    & 0.748     & \textbf{0.825}     & 0.817      \\
    miniboone   & 0.925     & \textbf{0.929}     & 0.894      \\
    \bottomrule
  \end{tabular}

  \caption{Results in hard budget settings. The \emph{rl-$\lambda$} is the performance in the average budget task and is plot only for comparison between the two tasks (to \emph{rl-hard}). The table shows the normalized area under the trade-off curve metric.}
  \label{fig:exp_hard_budget}
\end{figure}

In hard budget setting, we compare to recent heuristic-driven approach by \citet{kachuee2019opportunistic}, called Opportunistic Learning (\emph{oplearn}). In this algorithm, an auxiliary reward is defined as a change in prediction uncertainty, when some feature is added. Two separate networks are trained -- one estimating class probabilities, the other predicting the auxiliary reward. During test-time, the features are greedily acquired according to the predicted reward, and classification is made when the target budget is reached. As the method stands, it is hard to compare the used heuristic with the CwCF objective, and only experimental results show that it may be a good idea to follow this approach. Also, if only an immediate reward is predicted ($\gamma = 0$), the model loses the capacity to predict into the future. Nevertheless, the reported performance was impressive, hence we selected the method for comparison. We do not compare to the work of \citet{kapoor2005learning}, since it solves a slightly different problem. In their case, they don't use a per-sample budget, but rather a budget for the whole training process.

We selected miniboone and diabetes datasets for experiments, because miniboone is easily evaluated and diabetes was used in the evaluation of Opportunistic Learning. The results can be seen in \fig{exp_hard_budget}, where the RL algorithm with hard budget setting is named \emph{rl-hard}. For comparison reasons, we also plot the performance on the average budget task (\emph{rl-$\lambda$}). This is to compare the \emph{tasks} themselves, not the algorithms.

Compared to average budget setting, the hard budget algorithm achieves lower performance in miniboone. This is expected, because in contrast to the hard budget setting, the average budget method \emph{can exceed} the target budget for selected samples. In diabetes, the performance of the hard budget method is better for a range of costs, which we attribute to overfitting of the average budget method. 

Compared to the Opportunistic Learning algorithm, our method achieves substantially better performance on both datasets. We attribute the result to the fact that RL method optimizes for the actual objective, while Opportunistic Learning method optimizes a heuristic objective, which may not be optimal in the first place. Note that we tried to use a comparable number of parameters in the Opportunistic Learning algorithm, but observed decreased performance.

We see a similar effect as in the average budget settings -- if the increased budged does not result in increased accuracy, the model learns to stop acquiring features prematurely, to save resources. Note that in the hard setting, RL \emph{never} exceeds the specified budget.

\subsection{Missing features}
In these experiments we assume that there is a sparse training set, while during the test time, the models can select any feature. That corresponds to the case of the mentioned medical domain, where past data can be elevated for training. However, it is difficult to obtain a real dataset with these attributes and therefore we decided to create a custom synthetic dataset. We artificially drop some percentage of features from the miniboone dataset. We created four versions, with 25\%, 50\%, 75\% and 90\% of features missing. The syntetic datasets were created with an assumption that the features are missing completely at random (MCAR) and the fact that a feature is missing has no predictive power. 

\begin{figure}[t]
  \centering
  {
  \setlength{\tabcolsep}{0pt}
  \begin{tabular}{cccc}
    \includegraphics[width=0.45\linewidth]{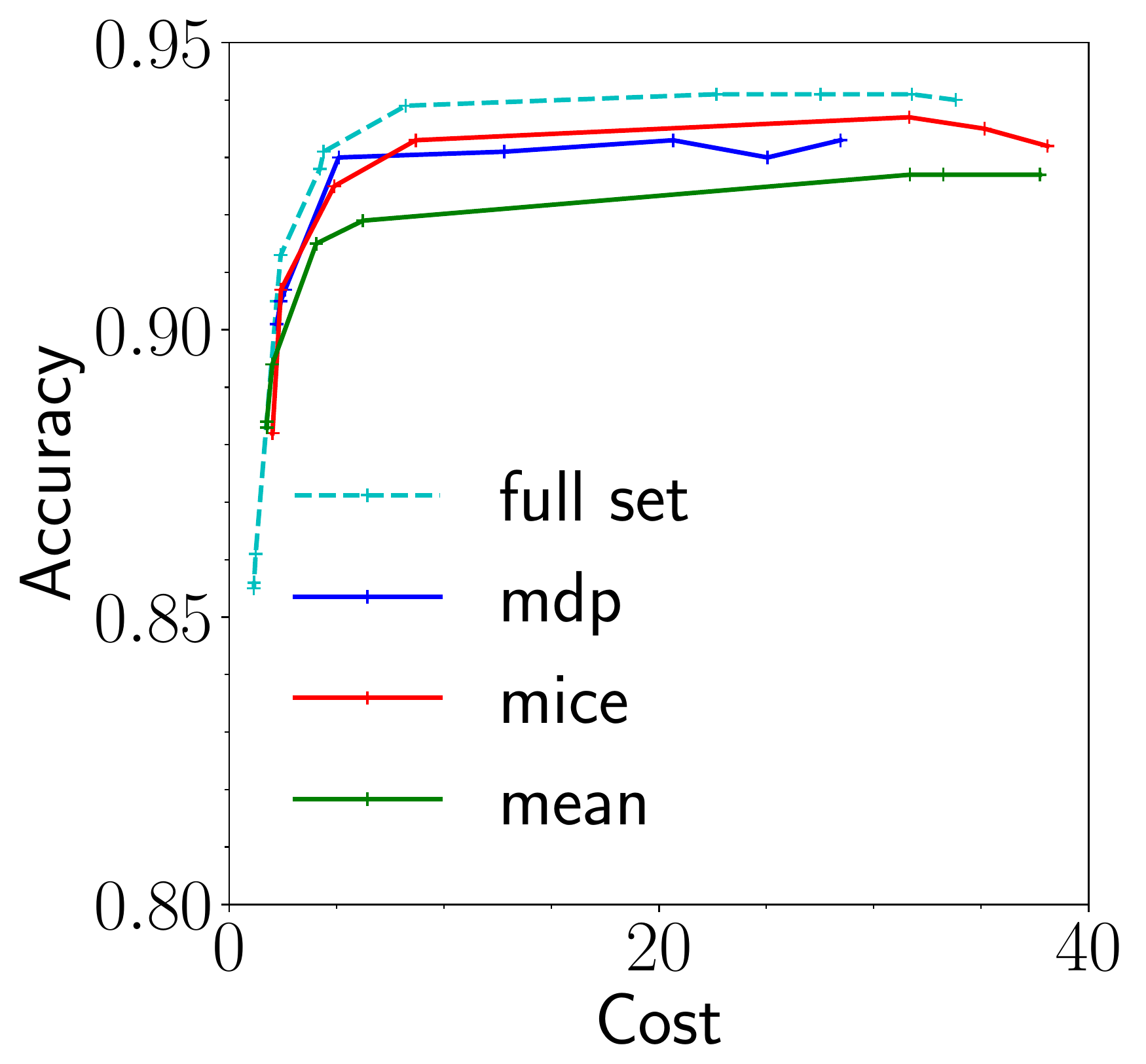} &
    \includegraphics[width=0.45\linewidth]{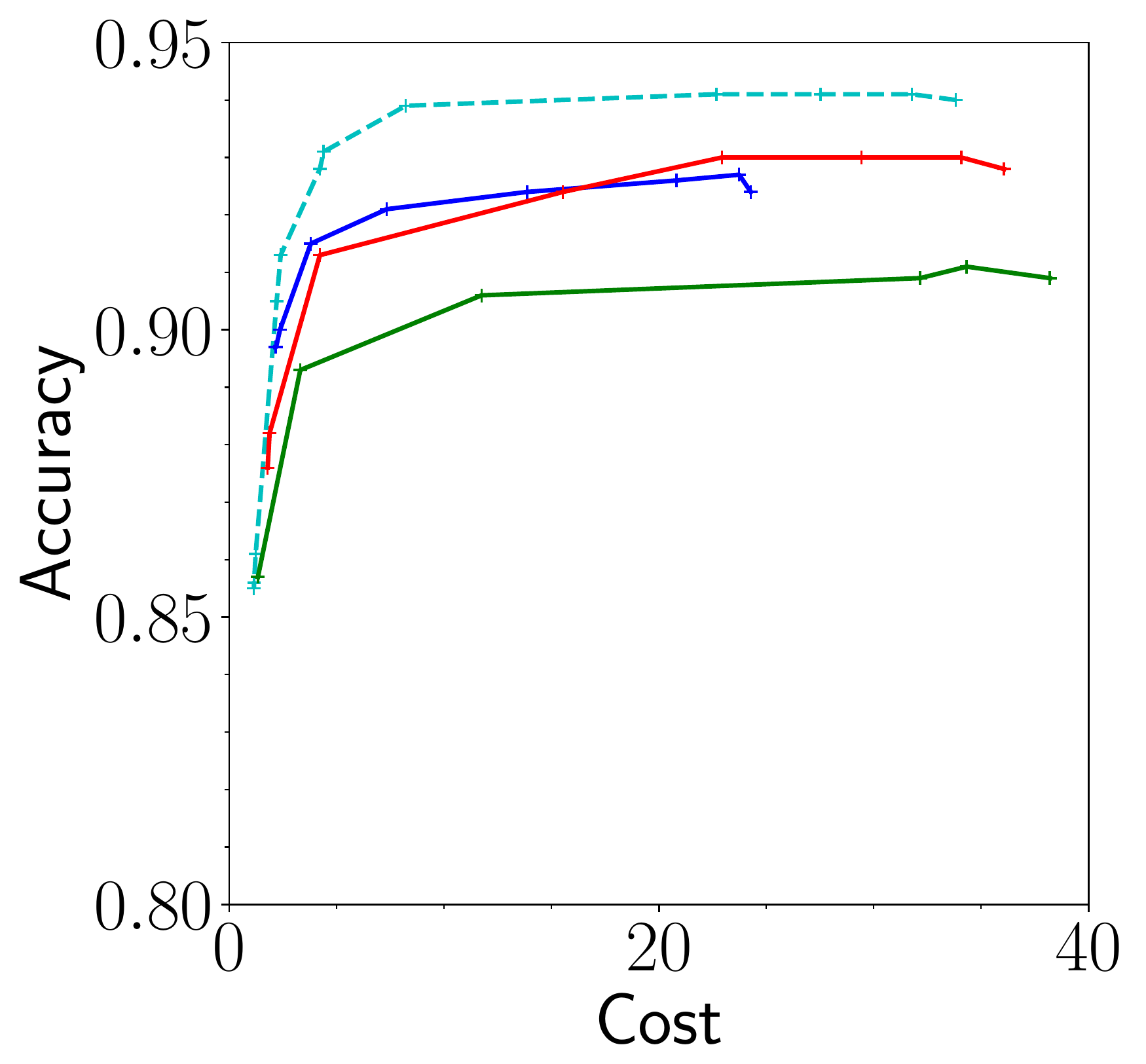} \\
    (a) 25\% & (b) 50\% \\

    \includegraphics[width=0.45\linewidth]{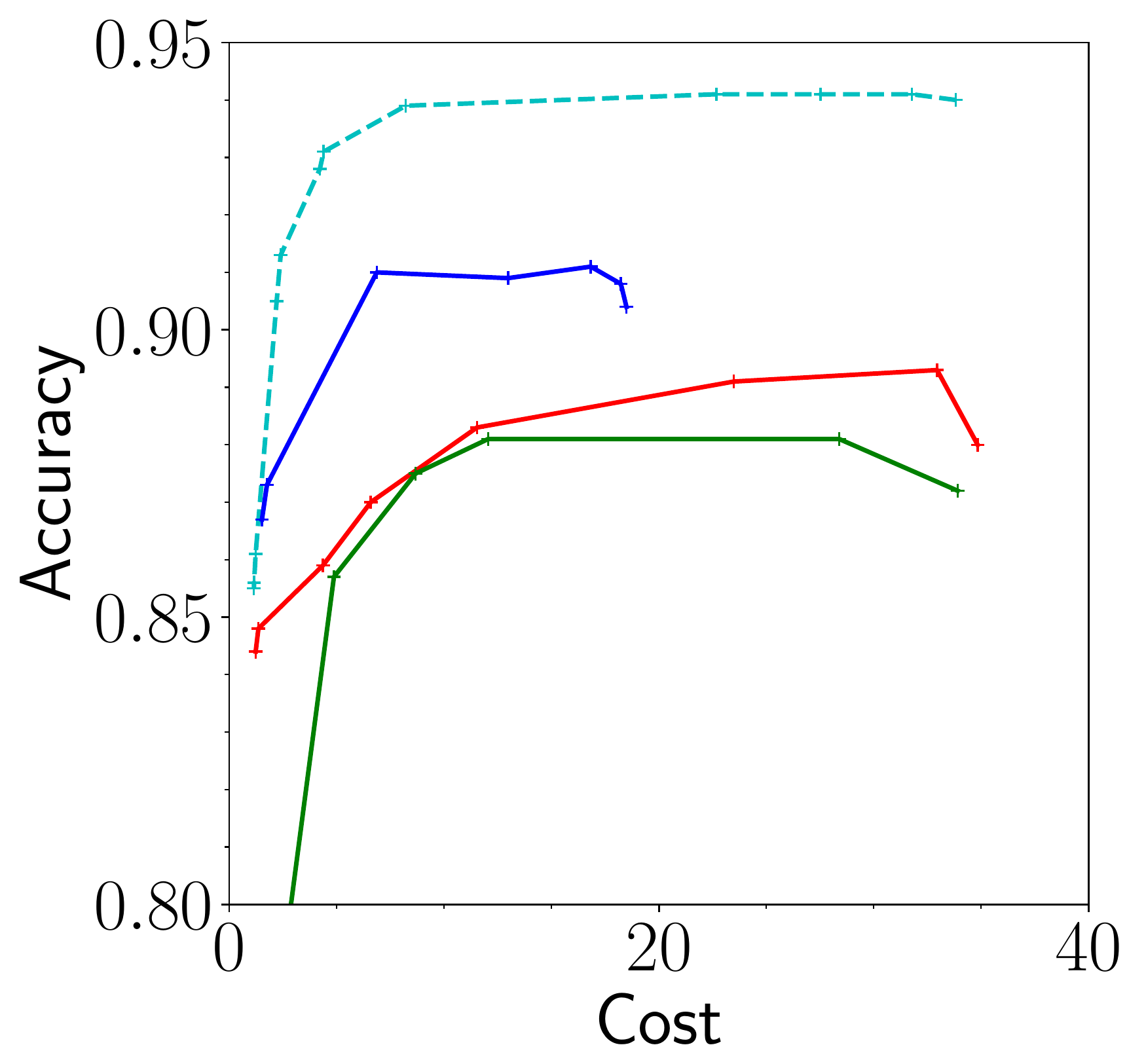} &
    \includegraphics[width=0.45\linewidth]{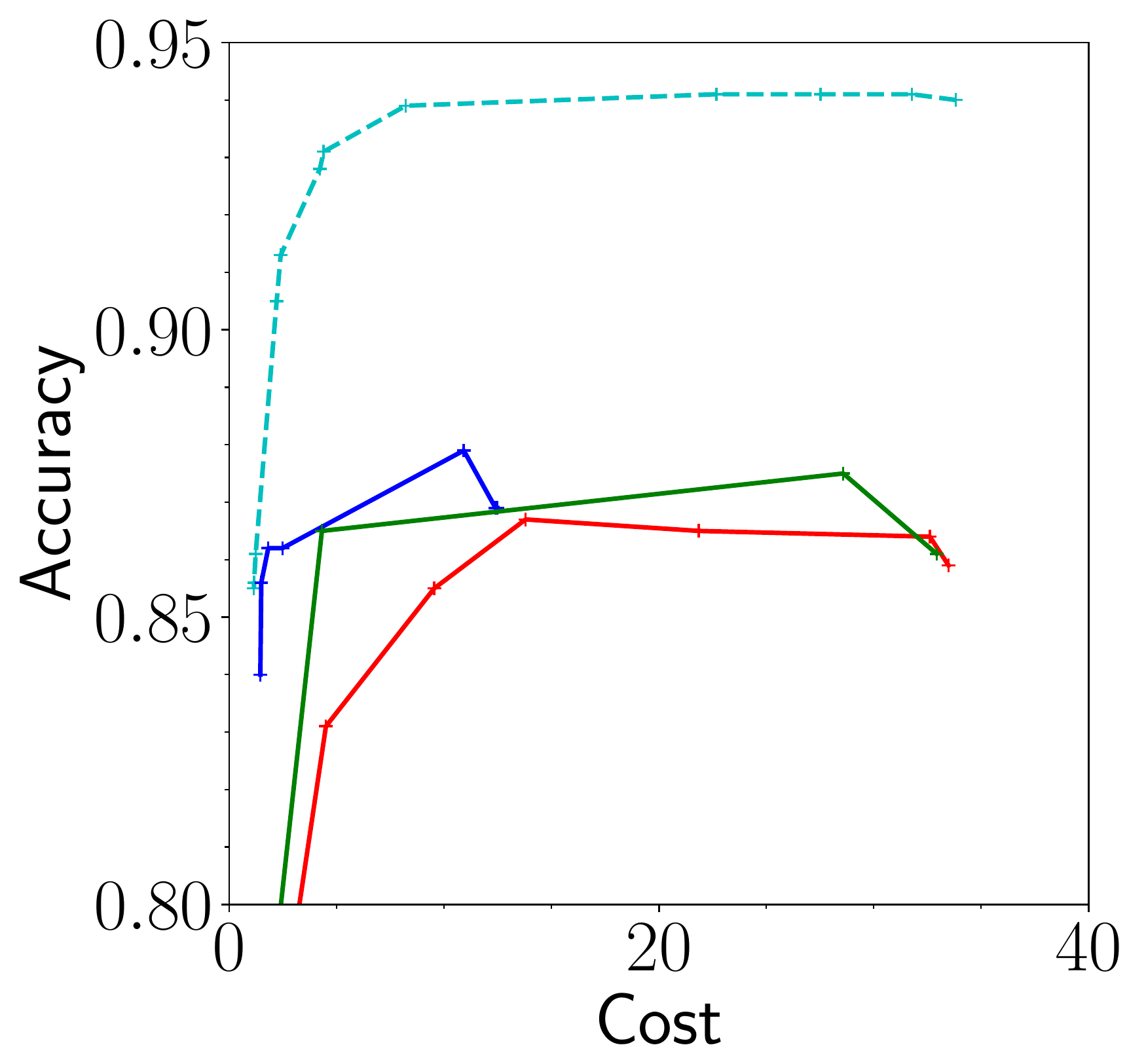} \\
    (c) 75\% & (d) 90\% \\
  \end{tabular}
  }
  \vspace*{2mm}

  \begin{tabular}{lrrr}
    \toprule
    Dataset    & mean      & mice      & mdp    \\
    \midrule
    miniboone-25      & 0.919     & \textbf{0.928}     & 0.926      \\
    miniboone-50      & 0.903     & \textbf{0.921}     & 0.919      \\
    miniboone-75      & 0.871     & 0.882     & \textbf{0.904}      \\
    miniboone-90      & 0.865     & 0.854     & \textbf{0.874}      \\
    \bottomrule
  \end{tabular}

  \caption{The tested methods were trained with sparse miniboone dataset, where the stated percentage of features is missing. We compare performance with training on the \emph{full set} without missing features, the altered \emph{mdp} method and \emph{mice} and \emph{mean} imputation algorithms. Trained models were evaluated on the complete dataset with all features. The table shows the normalized area under the trade-off curve. Numbers after the dataset name identifies the percentage of missing features during training.}
  \label{fig:exp_missing}
\end{figure}

We implement the method described in Section\;\ref{sec:method_missing} (\emph{mdp}). We use two baseline methods -- first, we simply impute the missing features with their \emph{mean} and train the usual way. Second, we use \emph{MICE} algorithm \citep{azur2011multiple}, which assumes linear dependencies between features. It works by iteratively predicting missing values with a linear regression over known or already predicted features and repeating this process several times. The imputed dataset is then regarded as complete and we train our method in a standard way. For comparison reasons, we also plot the performance on the \emph{full set} without any missing features.

In \fig{exp_missing} we present the results. We see incremental degradation of performance when an increasingly larger percentage of features is missing. The results show that the version with altered MDP performs robustly well. It performs comparably when less than 50\% of the features are missing, and performs substantially better with sparser datasets. The mdp method does not involve any preparation and can be directly used in any sparse dataset. It also highlights the flexibility of the RL method. In the case of the MICE imputation method, it has to be noted that the preparation process takes a non-negligible amount of time (about 15 minutes in the miniboone dataset).

\section{Related work} \label{sec:related_work}
Classification with Costly Features problem has been approached from many directions, with many different types of algorithms. But to our knowledge, there is no single framework that can work with both average and hard budgets, is flexible and perform robustly as our method. 

Most closest works to this article are \citep{dulac2011datum}, which used Q-learning with limited linear regression, resulting in inferior performance. Recent works \citep{janisch2019classification,shim2018joint} replace the linear approximation with neural networks and report superior performance. However, these methods focus only on the average budget problem and introduce an unintuitive trade-off parameter $\lambda$. In \citep{janisch2019classification} the authors showcase the flexibility of the network by incorporating an external classifier as a separate feature.

Following works focus only on the average budget problem. \citet{contardo2016recurrent} use a recurrent neural network that uses attention to select blocks of features and classifies after a fixed number of steps. There is also a plethora of tree-based algorithms \citep{xu2012greedy,kusner2014feature,xu2013cost,xu2014classifier,nan2015feature,nan2016pruning,nan2017adaptive}. 

A different set of algorithms employed Linear Programming (LP) to this domain \citep{wang2014lp,wang2014model}. \citet{wang2014model} use LP to select a model with the best accuracy and lowest cost, from a set of pre-trained models, all of which use a different set of features. The algorithm also chooses a model based on the complexity of the sample.

\citet{wang2015efficient} propose to reduce the problem by finding different disjoint subsets of features, that are used together as macro-features. These macro-features form a graph, which is solved with dynamic programming. In large problems, the algorithm can be used to find efficient groupings of features which would then be used in our method.

\citet{trapeznikov2013supervised} use a fixed order of features to reveal, with increasingly complex models that can use them. However, the order of features is not computed, and it is assumed that it is set manually. Our algorithm is not restricted to a fixed order of features (for each sample it can choose a completely different subset), and it can also find their significance automatically.

Recent work \citep{maliah2018mdp} focuses on CwCF with misclassification costs, constructs decision trees over features subsets and use their leaves to form states of an MDP. They directly solve the MDP with value-iteration for small datasets with the number of features ranging from 4-17. On the other hand, our method can be used to find an approximate solution to much larger datasets. In this work, we do not account for misclassification costs, but they could be easily incorporated into the rewards for classification actions.

\citet{tan1993cost} analyzes a problem similar to our definition, but algorithms introduced there require memorization of all training examples, which is not scalable in many domains.

The hard budget case was explored in \citep{kapoor2005learning}, who studied random and heuristic based methods. \citet{deng2007bandit} used techniques from the multi-armed bandit problem. There are also theoretical works \citep{cesa2011efficient,zolghadr2013online}. \citet{kachuee2019opportunistic} crafted a heuristic reward and used RL to maximize it.

\section{Conclusion} \label{sec:conclusion}
In this work, we presented a flexible reinforcement learning (RL) framework for solving the Classification with Costly Features (CwCF) problem. We build on an established work, that already showcased the superior performance of RL in this problem. We modified it to work with a directly specified budget in average and hard budget cases. For the average case, we introduced the Lagrangian theory to automatically find suitable parameters. We also modified the framework in a principled way, to be able to work with datasets with missing features. All settings were evaluated on several diverse datasets and we report that our method robustly outperforms other algorithms in most settings. 

The flexibility of the RL framework was successfully demonstrated by all mentioned versions of the algorithm. We showcased its robustness (it performs well across all datasets) and the ease of use (almost all hyperparameters stay the same across all datasets and algorithm variations). Also, we note that, being a standard RL algorithm, the method can benefit from any improvement in the RL area itself.

\section*{Acknowledgements}
We thank Cheng Zhang for suggesting the baseline method. This research was supported by the European Office of Aerospace Research and Development (grant no. FA9550-18-1-7008) and by The Czech Science Foundation (grants no. 18-21409S and 18-27483Y). The GPU used in this research was donated by the NVIDIA Corporation. Computational resources were provided by the CESNET LM2015042 and the CERIT Scientific Cloud LM2015085, provided under the program Projects of Large Research, Development, and Innovations Infrastructures.

\bibliographystyle{spbasic}
\bibliography{paper}

\clearpage
\section*{Appendix}
\setcounter{section}{1}
\renewcommand\thetable{\Alph{section}.\arabic{table}}
\begin{table}[h]
    \centering
    \begin{tabular}{llr}
      \toprule
      Symbol                          & description                                               & value               \\
      \midrule
      $|\mathcal{E}|$                 & number of parallel environments                           & 1000                 \\
                                      & maximum number of steps                                   & $100 \times \text{ep\_len}$ \\
      $\gamma$                        & discount-factor                                           & 1.0                 \\
      Retrace-$\lambda$               & Retrace parameter $\lambda$                               & 1.0                 \\
      $\rho$                          & target network update factor                              & 0.1                 \\
      $|\mathcal B|$                  & number of steps in batch                                  & 50k                 \\
      $|\mathcal M|$                  & number of episodes in memory                              & 40k                 \\
      $\epsilon_\text{-start}$      & starting exploration                                    & 1.0                 \\
      $\epsilon_\text{-end}$        & final exploration                                       & 0.1                 \\
      $\eta_\text{-start}$        & starting $\eta$-greediness of target policy $\pi$   & 0.5                 \\
      $\eta_\text{-end}$          & final $\eta$-greediness of target policy $\pi$    & 0.0                 \\
      $\epsilon_\text{steps}$         & length of exploration phase                               & $2 \times \text{ep\_len}$ \\
      LR-pretrain                     & pre-training learning-rate                                & $1 \times 10^{-3}$\\
      LR-start                        & initial learning-rate                                     & $5 \times 10^{-4}$\\
      LR-min                          & minimal learning-rate                                     & $5 \times 10^{-7}$\\
      LR-scale                        & learning-rate multiplicator                               & 0.5                 \\
      \bottomrule
    \end{tabular}

    \vspace*{2mm}
    (a) Global parameters
    \vspace*{4mm}

    \begin{threeparttable}
      \begin{tabular}{lrrl}
        \toprule
        Dataset                 & model size & ep\_len   & specific \\
        \midrule                                
        mnist\tnote{\dag}       & 512       & 10k       & $|\mathcal M|=10\text{k}, \text{LR-pretrain}=2 \times 10^{-5}, \text{LR-start}=10^{-5}$   \\
        cifar\tnote{\dag}       & 512       & 10k       & $|\mathcal M|=10\text{k}, \text{LR-pretrain}=2 \times 10^{-5}, \text{LR-start}=10^{-5}$   \\
        forest                  & 256       & 10k       &          \\
        miniboone               & 128       & 1k        &          \\
        diabetes                & 128       & 100       &          \\
        \bottomrule
      \end{tabular}
      \begin{tablenotes}\footnotesize
        \item[\dag] The replay size is lowered due to memory constraints.
      \end{tablenotes}
    \end{threeparttable}
    
    \vspace*{2mm}
    (b) Dataset parameters
    \vspace*{4mm}

    \caption{Algorithm hyperparameters. (a) All algorithm-level parameters. (b) Dataset-specific parameters. For each dataset, the \emph{ep\_len} (from epoch length) specifies a difficulty of a dataset and some other parameters are derived from it.}
    \label{tab:parameters}    
\end{table}

\end{document}